%% file: main.tex
\documentclass[twoside,leqno,twocolumn]{article}

\usepackage[letterpaper]{geometry}

\usepackage{ltexpprt}
\usepackage{algorithm,algorithmic,amsmath}
\usepackage{multirow}
\usepackage{graphics}
\usepackage{float}
\usepackage[caption = false]{subfig}
\usepackage[demo]{graphicx}
\usepackage{tikz}
\usepackage{xcolor}
\usepackage{balance}

\DeclareMathOperator*{\argmax}{arg\,max}
\DeclareMathOperator*{\BEE}{\mathbf{E}} 

\newcommand{\bx}{\mathbf{x}}
\newcommand{\bw}{\mathbf{w}}
\newcommand{\bb}{\mathbf{b}}
\newcommand{\bp}{\mathbf{p}}
\newcommand{\bq}{\mathbf{q}}
\newcommand{\bt}{\mathbf{t}}
\newcommand{\br}{\mathbf{r}}
\newcommand{\bc}{\mathbf{c}}
\newcommand{\ba}{\mathbf{a}}
\newcommand{\BE}{\mathbf{E}}
\newcommand{\CM}{\mathcal{M}}
\newcommand{\CS}{\mathcal{S}}
\newcommand{\CU}{\mathcal{U}}

\newcommand{\CA}{\mathcal{A}}
\newcommand{\CO}{\mathcal{O}}

\newcommand{\CT}{\mathcal{T}}
\newcommand{\CD}{\mathcal{D}}
\newcommand{\CR}{\mathcal{R}}
\newcommand{\CW}{\mathcal{W}}
\newcommand{\CN}{\mathcal{N}}
\newcommand{\CL}{\mathcal{L}}
\newcommand{\bs}{\mathbf{s}}

\newcommand{\cpmtest}{\mbox{cpm}_{\mbox{te}}}
\newcommand{\cpmtrain}{\mbox{cpm}_{\mbox{tr}}}

\usepackage{amsthm}
\theoremstyle{definition}
\newtheorem{assumption}{Assumption}
\newtheorem{remark}{Remark}

\newcommand{\gl}[1]{\textcolor{red}{}}
\newcommand{\sm}[1]{\textcolor{blue}{}}
\renewcommand{\Pr}{\mbox{P}}
\begin{document}

\title{\Large Optimal Bidding Strategy without Exploration  in Real-time Bidding}
\author{Aritra Ghosh \thanks{UMass, Amherst, MA, USA}\\ {\small \texttt{arighosh@cs.umass.edu}}
\and Saayan Mitra \thanks{Adobe Research, San Jose, CA, USA} \\ {\small\texttt{smitra@adobe.com} } \and Somdeb Sarkhel \footnotemark[2] \\ {\small\texttt{sarkhel@adobe.com}} \and Viswanathan Swaminathan \footnotemark[2]\\{\small\texttt{vishy@adobe.com}}}

\date{}

\maketitle


\fancyfoot[R]{\scriptsize{Copyright \textcopyright\ 2020 by SIAM\\
Unauthorized reproduction of this article is prohibited}}





\input{abs}

\input{intro}

\input{related}

\input{problem}

\input{method}
\input{exp}

\input{con}
\bibliographystyle{abbrv}
\bibliography{bid_irl}
\input{supplementary}

\end{document}

%% file: abs.tex
\begin{abstract} \small\baselineskip=9pt 
     Maximizing utility with a budget constraint is the primary goal for advertisers in real-time bidding (RTB) systems. The policy maximizing the utility is referred to as the optimal bidding strategy. Earlier works on optimal bidding strategy apply model-based batch reinforcement learning methods which can not generalize to unknown budget and time constraint. Further, the advertiser observes a censored market price which makes direct evaluation infeasible on batch test datasets. Previous works ignore the losing auctions to alleviate the difficulty with censored states; thus significantly modifying the test distribution. We address the challenge of lacking a clear evaluation procedure as well as the error propagated through batch reinforcement learning methods in RTB systems.  We exploit two conditional independence structures in the sequential bidding process that allow us to propose a novel practical framework using the maximum entropy principle to imitate the behavior of the true distribution observed in real-time traffic.  Moreover, the framework allows us to train a model that can generalize to the unseen budget conditions than limit only to those observed in history. We compare our methods on two real-world RTB datasets with several baselines and demonstrate significantly improved performance under various budget settings.
\end{abstract}

%% file: intro.tex
\section{Introduction}
Real-time bidding (RTB) has become the dominant mechanism for online advertising in current times.  One of the key challenges for the advertiser in the RTB system is to devise a sequential bidding strategy for bid requests to maximize some utility (e.g., impression, clicks, etc.) under some {budget constraint}.
Under repeated auctions with budget constraints, bidding the true value is not the optimal action in second-price auctions. We refer to the policy maximizing expected utility under a budget constraint as the \emph{optimal bidding strategy}.

In recent times, reinforcement learning (RL) achieves almost human-level performance in many games and control problems \cite{human}. 
{Reinforcement learning to bid ({RLB})}, a model-based RL approach for optimal bidding in RTB has been proposed recently \cite{rlb}.
Although {RLB} improves upon existing methods, {RLB} suffers from scalability and efficiency issues in RTB systems using a misspecified model learned from inadequate interactions. A better alternative is to use a model-free RL algorithm which requires a large number of interactions with the environment  (users, other participants in RTB) to learn the optimal strategy. However, in real-time bidding systems, learning the optimal bidding strategy is often a batch process with limited opportunity to interact with the environment. From historical interactions, the advertiser needs to learn the optimal action (bidding price). In reinforcement learning literature, learning optimal policy from historical interactions without exploration is often termed as \emph{batch reinforcement learning} \cite{batchrl-book}. 

Although model-free online RL algorithms are conceptually appealing, real-time bidding systems often have the choice of batch reinforcement learning. Most early approaches of optimal bidding strategy can be classified as some form of {batch RL} \cite{rlb,amin,ads-pomdp,linbid}. However, the single drawback of any {batch RL} algorithm is the limited generalization ability to new state space due to extrapolation error as shown lately in \cite{batch-rl-iclml}. Without exploration, often it is not feasible to extrapolate under new budget constraints.
Besides, a clear evaluation procedure is lacking in the literature of online advertising due to the nature of the batch data.
Without interacting with the environment, it is not viable to evaluate reward for any action other than the one taken in the test batch dataset. Further, in the context of RTB, the behavior of the market is censored when the advertiser loses an auction; hence the advertiser often observes censored states which form the test dataset.  Early methods of optimal bidding strategy remove the censored/unobserved part of the test dataset for evaluation leading to a different distribution than observed in real-time environments. The dataset shift in the true distribution and the truncated train-test distribution adopted in previous research makes such models unsuitable for directly applying to the real-world traffic.

The batch framework is particularly important as the dataset is usually collected from an (unknown) off-policy strategy in the RTB system. Further, conducting real-traffic A/B testing is not often feasible and cost-effective in practice.
We address the deficiencies in the batch RL methods such as lack of a clear evaluation procedure as well as the error propagation through the training methods in RTB systems. Using two conditional independence structures in the sequential bidding process, we propose a novel framework applying the maximum entropy principle to imitate the behavior of the true distribution observed in real-time traffic. The simulated environment allows accurate evaluation of any model in the RTB system without the dataset shift.
Further, the framework enables training any model-free RL algorithm that can generalize to an unseen budget and time constraints, beyond the states observed in {historical interactions}. We compare our methods on real-world public RTB data-sets with several baselines and show that our framework significantly improves generalization performance under various budget settings.

%% file: related.tex
\vspace{-.1in}
\section{ Background and Related Work}
\paragraph{Reinforcement Learning.}
A Markov Decision Process (MDP) $\CM$ is represented as a tuple $(\CS, \CA,\CT, \CR, \gamma)$ which consists of a set of states $s\in \CS$, a set of actions $a\in \CA$, a transition function $\CT(s,a,s')= \Pr(s_{i+1}=s'|s_i=s,a_i=a)$, a reward function $\CR(s,a)$, and, a discount parameter $\gamma$ \cite{rlbook}. $s_i,a_i,r_i$ are the state, action, and, reward at time step $i$ respectively.  A policy is defined as $\pi: \CS\times \CA \rightarrow [0,1]$ representing the conditional distribution over actions given the state that the agent follows. The goal of the agent is to find the policy $\pi$ that maximizes the expected discounted reward over the episode \cite{rlbook}.
State-Action value function is the expected reward  that can be obtained following policy $\pi$  starting from a state-action pair $(s,a)\in \CS\times\CA$:
\begin{eqnarray*}
&Q^{\pi}(s,a) &= \BE [\sum_{\tau=t}^{\infty}    \gamma^{\tau-t}r_{\tau}|s_t =s, a_t =a,\pi            ]
\end{eqnarray*}
The optimal value function for the state-action pair, $Q^{\ast}(s,a) = \max_{\pi} Q^{\pi}(s,a)$, satisfies the Bellman Optimality equation \cite{rlbook},
$Q^{\ast}(s,a) = \BE  [\CR(s,a)+\gamma \sum_{s'}\CT(s,a,s')\max_{a'} Q^{\ast}(s',a') ]  .$ Similarly, value function from a state $\bs\in \CS$ is the expected reward from that state following policy $\pi$, $V^{\pi}(s) = \BE_{a\sim \pi} Q^{\pi}(s, a)$.

 \paragraph{Batch reinforcement learning.} In batch RL, the agent does not have opportunities to interact with the environment \cite{batchrl-book}.  If the policy used to collect the experiences is known, we can use the deep Q-network (DQN) with {importance sampling} (IS) to learn from the batch experiences (dataset) \cite{human}. However, DQN with IS suffers from high extrapolation error \cite{batch-rl-iclml}. Further, in an RTB system, the policy that gathered the historical data is usually unknown. A general approach to learn from batch data is to learn fitted Q-iteration (possibly with a deep network) \cite{fdqn,fqn}. Nevertheless, without exploration, the performance of neural fitted Q iteration is limited to the experiences gathered.
 
 \paragraph{Real-time Bidding.}
In RTB systems, ad display opportunities are traded using second-price auctions in real-time from the publishers (sellers) to the advertisers (buyers) through the ad-exchange. When a user visits a publisher's page, the supply-side platform, acting on behalf of the publisher, requests the ad-exchange for advertisements to fill up the vacant slots. The ad-exchange, in turn, announces the request to the demand-side platforms (DSP) who represent the advertisers. Subsequent to getting a bid request, DSP chooses one among several ads to bid and participates in the auction. The ad exchange picks the highest bidder and the winning DSP pays the second price (or market price). DSP observes the market price if it wins the auction; otherwise, in case of losing the auction, only the lower bound on the market price is known. This mixture of observed and partially observed data is known as \emph{censored data}. 
The purpose of DSP is to maximize some utility (impression, click, or, conversion) under some budget and time constraint for each of its advertisers. In the rest of the paper, we use the term {advertiser} and {DSP} interchangeably for simplicity.

We represent a bid request as $\bx$, market price as $\bw$ (can be unknown), budget left as $\bb$, and, time left as $\bt$. Advertiser bids $\ba$ for the bid request $\bx$ with cost $\bc$ ($\bw$ when the advertiser wins the auction and $0$ otherwise). The advertiser observes a utility reward $\br$ (impression, click, or, conversion). The objective of the advertiser is to maximize the total reward obtained given the budget and time constraint with a policy $\pi([\bx,\bb,\bt], \cdot)$:
   \[\max_{\ba \sim \pi}\sum^{\bt}_1 \BE_{\bx} [\br|\bx],\quad \mbox{such that}\quad \sum^{\bt}_1 \BE_{\bx}[\bc|\ba, \bx]\leq \bb.\]

\paragraph{Optimal Bidding Strategy.}
To compute the optimal bidding strategy, the advertiser needs to estimate the expected utility and the expected cost. Previous research extensively studied utility estimation problems (such as click-through rate estimation) \cite{ctr} and the bid landscape forecasting problem for computing expected cost \cite{cr,mdn-cr}.
In real-world applications, due to simplicity and scalability, the bidding systems usually employ a simple linear bidding strategy \cite{linbid} where the advertiser bids proportional to their expected value (utility). 
Early attempts for optimal bidding strategy include model-based MDP and partially observable MDP formulations \cite{amin,ads-pomdp} which works in the context of sponsored ads. In \cite{rlb}, the authors proposed the RLB model to solve the Bellman equation on a simplified model. Solving the Bellman equation has an attractive property of computing the optimal strategy when the model is correct. However, to tackle large continuous state and action space, the {RLB} model assumes the market price and the winning rate are independent of the bid requests. Moreover, inferring the optimal action has $\CO(k)$  time complexity and $\CO(T^2k)$ memory complexity for each bid request ($T,k$ is the maximum number of time steps left and the number of bid price respectively). This is unacceptable in the real world situation. {RLB} proposed to use a segmented time window with neural network approximation leading to further approximation besides the model simplification. Besides, model-based batch reinforcement learning algorithms do not have the opportunity to explore with the real environment making them prone to extrapolation errors \cite{batch-rl-iclml}.

 In this paper, we frame the problem as learning the optimal policy from batch datasets (with censored states) without the choice of explorations and access to the policy used to collect the dataset. 
 The second problem we consider is to learn a correct evaluation procedure from such offline datasets with censored states. Note that, for losing an auction, we only have a lower bound on the market price; consequently, it is not possible to accurately evaluate the reward when evaluating the batch-test dataset. Earlier research excludes all the bid requests where the market price is not known leading to a distributional shift.

%% file: problem.tex
\vspace{-0.1in}
\section{Problem Definition} 
\label{sec:problem}
We define the MDP $\CM$  for the optimal bidding strategy problem  $(\CS, \CA,\CT, \CR, \gamma)$ with policy $\pi$ as follows:

 \textbf{i)} We represent the {\bf state}  $\bs\in \CS$  at time step $i$ as $\bs_i = [\bx_i\oplus \bw_{i}\oplus \bb_{i}\oplus \bt_i]$  where $\bx_i, \bb_i, \bt_i$ are the bid request, budget left, and, time left respectively and $\oplus$ represents the concatenation operator. $\bw_{i}$ is the market price of  auction at time step $i$ and observed (might be censored) at time step $i+1$.
Thus we use another notation $\bs^{O}_i = [\bx_i\oplus \bb_i\oplus \bt_i]$ to represent the partially observed state without the market price $\bw_i$ at time $i$. 
The market price $\bw_i$ represent sufficient statistics about the market (consisting of many DSP) behavior overall. 
 \textbf{ii)} {\bf Action} $\ba_i\in \CA$ is the bid price for bid request $\bx_i$ at time step $i$. 
\textbf{iii)} {\bf Reward} $\br_i=\mbox{imp}_i \mbox{ or }  \mbox{click}_i$  is the user/market response for bid request $\bx_i$ and bid price $\ba_i$; when impression is the utility, reward is 1 if $\ba_i> \bw_i$ and 0 otherwise. When the utility of interest is click, besides the impression, the user needs to click the ad  for advertiser to  get reward 1.
\textbf{iv)} {\bf Discount Parameter} $\gamma$ is usually set as $1$ in RTB as the objective is to maximize total rewards with budget constraints. 
\textbf{v)} {\bf Policy} $\pi(\bs^{O}=[\bx\oplus\bb\oplus\bt],\ \ba)$ is the probability of bidding $\ba$ when the bid request, time left, budget left are $\bx$, $\bt$, $\bb$ respectively.
\textbf{ vi)} We decompose the state representation into the \textbf{market-specific state} ($\bs_i^M = [\bx_i\oplus \bw_{i}]$) and the \textbf{advertiser state} ($\bs_i^A = [ \bb_{i}\oplus \bt_i]$). The \textbf{transition function} is: \[\CT(\bs_{i}, \ba_i, \bs_{i+1}) = \Pr(\bs_{i+1}= [\bs_{i+1}^M\oplus \bs_{i+1}^A]\quad|\quad \bs_{i}, \ba_i                ).\]
 We denote the batch dataset as a collection of trajectories $\CD= \{\tau_i\}_{i=1}^n$  where trajectory $\tau_i = (\bs_1,\ba_1, \br_1, \bs_2,\ba_2,\br_2,\cdots, \bs_T,\ba_T,\br_T)$  is the sequence of state, action, reward in the $i^{\mbox{th}}$ episode.
We need to learn the optimal policy $\pi^{\ast}(\bs^{O}=[\bx_i\oplus \bb_i\oplus \bt_i],\cdot)$ from the batch dataset; further we need to devise an evaluation framework that represents the actual test distribution advertiser observes. Moreover, we are interested in policy $\pi$ that can generalize to any advertiser state space ($\bs^A = [ \bb\oplus \bt]$).
\begin{remark}
Before delving into methodology, we state two important conditional independence properties that we use throughout our formulation. \textbf{i)} Current market-specific state and advertiser state are conditionally independent given the last state and the last action. Thus the transition function simplifies to:
\begin{align}
\Pr(\bs_{i+1}|  \bs_{i},\ba_i) = \Pr( \bs_{i+1}^M| \bs_{i}, \ba_i                ) \cdot \Pr( \bs_{i+1}^A| \bs_{i}, \ba_i                )
\label{eq:1st}
\end{align}
We can consider the impact of a bidding action taken by the advertiser is negligible on the decision of future bid requests generated by large number of users as well as the bidding behavior of many other DSPs and advertisers. \textbf{ii)} Market-specific states at any two time points  $i\neq j$ are independent of each other. 
\begin{align}
\Pr(\bs_{i}^M = s_i,\bs_{j}^M=s_j) = \Pr(\bs_{i}^M=s_i) \Pr(\bs_{j}^M=s_j)
\label{eq:2nd}
\end{align}
We can assume numerous bid requests generated  from many users at different timesteps are independent of each other.
\label{remark:cond}
\end{remark}

%% file: method.tex
\vspace{-0.2in}
\section{Methodology}
\input{tikz_summary}   
In Figure~\ref{fig:summary}(a), we show the graphical model for the sequence of states (market and advertiser), actions and rewards.  The key idea is that we need to imitate the behavior of the environment from the batch dataset to evaluate the correct metric for any policy and to successfully learn the optimal policy starting with any advertiser state ($\bs^A = [ \bb\oplus \bt]$).

The two conditional independence in Remark~\ref{remark:cond} implies that the next state of the advertiser depends on the past state, and, action while the market-specific states are mutually independent at any two timesteps. Thus, for the agent to explore in a simulated environment, we only need the behavior of market-specific state trajectory $(\bs_1^M\rightarrow \bs_2^M \cdots\rightarrow \bs_n^M)$ and the reward distribution $\bp_r(\br; \ba,\bs^M)$. Using these models, the agent can instantiate with any starting budget $\bb$ and time constraints $\bt$ (that constitutes advertiser state) in the environment and explore to learn the optimal bidding strategy that can generalize to unknown advertiser state space. 
Further, we can evaluate any bidding strategy under any advertiser state space if we have access to the model that can imitate the market behavior and reward distributions. Thus the problem reduces to learning a market model that can sample close to the distribution used to generate the batch dataset. In Figure~\ref{fig:summary}(b), we show a schematic diagram of the training and testing frameworks that exploits the conditional structure on the graphical model to simulate the training and testing environment; details of which follow next.
Finally, we note that the market (users, other bidders) behaves rationally and maximize their long-term reward; often the user clicks ads only if it is relevant to them and other bidders optimize their cost rationally \cite{usergan}.  We explicitly state this assumption.
\vspace{-0.06in}
\begin{assumption}
\label{ass:optimal}
   The market optimizes some unknown cost function and is near-optimal in making decisions.
   \vspace{-0.07in}
\end{assumption}
\subsection{Maximum Entropy Market Model}
In inverse reinforcement learning (IRL),  the task is to learn the latent cost function as well as the optimal policy from expert's trajectories \cite{apprenticeship,maxirl}. To imitate the behavior of the market, we need to learn the state dynamics and reward distribution of the market. 
The (near) optimal trajectories from market are $\{\tau_i^M\}_{i=1}^n$ where $\tau_i^M = (\bs_1^M,\br_1, \bs_2^M,\br_2,\cdots,\bs_T^M,\br_T)$, $\bs^M$ is the the market-specific state and $\br_i$ is the observed reward by  the advertiser. The state consists of bid request $\bx_i$ in addition to the (censored) market price $\bw_i$ while the reward $\br_i$ is the observed ad impression/click.

For simplicity of derivation, we introduce a separate notation for the maximum entropy market model with a new MDP $\CM^M$ along with a modified state and action space.
For the market MDP $\CM^M$, we denote $\bx_i$ as the new \textbf{market state} and $\ba_i^M = [\bw_i\oplus \br_i]$ as the \textbf{market action}.  The decomposition follows our intuition that the market action (market price, click behavior of user) is (near) optimal. Finally, note that we do not observe $\bw_i,\br_i$ always; in case of losing auction only a lower bound of the market price $\bw_i$ is known which we handle later using censored regression-based approaches \cite{cr}. Under the new notation, we need to model the market trajectories $\{\tau_i^M\}^n_{1}$ where $\tau_i^M = (\bx_1, \ba^M_1, \cdots, \bx_T,\ba_{T}^M)$.

We formulate the market trajectory distribution as, $\Pr_{\phi}(\tau^M) = \frac{\exp(-c_{\phi}(\tau^M))}{Z}$ where $c_{\phi}(\tau^M) = \sum_{i=1}^n c_{\phi}(\bx_i,\ba_i^M)$ is the latent cost function the market is optimizing with parameter $\phi$, and, $Z$ is the partition function. This form follows our assumption that market is near-optimal as trajectories with lower cost are  exponentially more probable than the trajectories with higher cost. Further, the model prefers all the  trajectories with the same latent cost functions, equally; thus maximizing the entropy of the distribution \cite{maxirl}. 
We maximize log-likelihood to optimize $\phi$ such that historical market trajectories have a low-cost value:
\[    \ell(\phi)= \BE_{\tau\sim \CM^M } \log \Pr_{\phi}(\tau)= \BE_{\tau\sim \CM^M } [-c_{\phi}(\tau) -\log Z] \]
We further simplify with Eq.~\ref{eq:2nd} that $(\bx_i,\ba_i^M)\perp (\bx_j,\ba_j^M)$ (note we decompose the market-specific state $\bs^M$ into $\bx, \bw$ in the market MDP formulation). Thus a random state-action pair of the market trajectory will have the same expected values as the whole trajectory. Furthermore, we decompose the  cost function as $c_{\phi}(\bx, \ba^M)  =c^1_{\phi}(\bx) +c^2_{\phi} (\ba^M;\bx)$.
\begin{eqnarray}
    &\ell(\phi)&= \BE_{(\bx, \ba^M)\sim \bs^M } \log \Pr_{\phi}(\bx,\ba^M) \nonumber \\
    && = \underbrace{\BE_{\bx\sim \bs^M } \log \Pr_{\phi}(\bx)}_\text{Market State Obj.} +  \underbrace{\BE_{\ba^M\sim \bs^M|\bx } \log \Pr_{\phi}(\ba^M|\bx)}_\text{Market Action Obj.} \label{eq:main}\\
    && = \quad\quad \quad \ell_1(\phi)\quad\quad +\quad\quad\quad \ell_2(\phi)\nonumber
\end{eqnarray}
\paragraph{Market State Model.} We start with the first term $\ell_1(\phi)$ for the state $\bx$ of the MDP $\CM^M$.
\begin{eqnarray*}
  &  \ell_1(\phi)    &= \BE_{\bx\sim P_r } [-c^1_{\phi}(\bx) ] -\log Z_1\\
&& = \BE_{\bx\sim P_r } [-c^1_{\phi}(\bx) ] -\log [\BE_{\bx\sim \bq_{\theta}}\frac{\exp(-c^1_{\phi}(\bx))}{\bq_{\theta}(\bx)}]
\end{eqnarray*}
where we denote $P_r$ as the true distribution of bid requests $\bx$.
Computing partition function $Z_1= \int_x \exp(-c^1_{\phi}(\bx))dx$ is intractable; thus we compute $Z_1$ with samples from a parameterized distribution $\bq_{\theta}$ with parameter $\theta$ and multiply with importance sampling weights  to get a consistent estimate of $Z_1$. We use a single Monte-carlo sample to compute the unbiased estimate of the  partition function $Z_1$. The objective becomes:
\begin{eqnarray*}
\ell_1(\phi)& \simeq \BE_{\bx\sim P_r } [-c^1_{\phi}(\bx) ] -\log [\frac{\exp(-c^1_{\phi}(\bx))}{\bq_{\theta}(\bx)}]_{\bx\sim \bq_{\theta}}\\
& = \BE_{\bx\sim P_r } [-c^1_{\phi}(\bx) ] - [-c^1_{\phi}(\bx))]_{\bx\sim \bq_{\theta}}+\mbox{const}
\end{eqnarray*}
where the constant term depends only on sampler parameter ${\theta}$ (not $\phi$).
However, sampling from any distribution $\bq_{\theta}$ with importance sampling have a high variance in the estimate of $Z_1$.  The optimal distribution (with least variance) to sample from is $\bq^{\ast}_{\theta}(\bx)\propto \exp(-c^1_{\phi}(\bx))$. Thus similar to guided cost learning \cite{gcl}, we optimize $\bq_{\theta}$ as well to make it more likely distribution under the market cost function ${\phi}$. The objective for the sampler $\bq_{\theta}$ is:
\[L_1(\theta) =\BE_{\bx\sim \bq_{\theta}}[-c^1_{\phi}(\bx)]\]
 The market sampler (we call it the generator) samples from $\bx \sim \bq_{\theta}$ to minimize the  cost on sampled trajectories while  the cost function parameter (we call it the critic) uses network parameter $c_{\phi}$ to maximize the cost for sampled trajectories and minimize the cost from the real trajectories.
The objective for market state model has the Wasserstein-1 distance form \cite{wgan,wgan-gp}:
\[W(P_r, q_{\theta}) = \sup_{||c_{\phi}||_L\leq 1} \BE_{\bx\sim P_r}[c_{\phi}(\bx)]-\BE_{\bx \sim \bq_{\theta}}[c_{\phi}(\bx)]    \]
where the supremum is over all the 1-Lipschitz functions $c_{\phi}:X\rightarrow R$. 
The
Wasserstein-1 distance with gradient penalty  mitigates the vanishing
gradient problem observed when minimizing Jensen-Shannon
divergence and empirically performs better \cite{gan,wgan,wgan-gp}.  Thus, similar to \cite{wgan-gp}, we optimize  Wasserstein distance with the gradient penalty term to learn the \textit{market state model}:
\[\min_{c_{\phi}} \max_{\bq_{\theta}}\BE_{\bx\sim P_r}[c_{\phi}(\bx)]-\BE_{\bx \sim \bq_{\theta}}[c_{\phi}(\bx)]+ \lambda     L_{gp}        \]
where $L_{gp}=\BE_{\hat{\bx}\sim P_{\hat{\bx}}} [||\nabla_{\hat{\bx}} c_{\phi}(\hat{\bx})||_2 -1)^2$ and $P_{\hat{\bx}}$ is the sampling distribution\footnote{$t\bx+(1-t)\Tilde{\bx}=\hat{\bx}\sim P_{\hat{x}} $ with $t\in \CU(0,1)$, $\bx\sim P_r$, $\Tilde{\bx}\sim \bq_{\theta}$.}. If we parameterize the cost function $c_{\phi}$ with a binary classifier, we get the standard generative adversarial network resulting in generative adversarial imitation learning \cite{gan,gail}.

\paragraph{Sampling State.} The market state $\bx$  or the bid request is multi-categorical (or binary) in nature. The generator (market sampler $\bq_{\theta}$) maps random vectors $z\in R^z$ (often multivariate standard Gaussian) to  generated inputs as $\hat{\bx} = G(z,\theta)$. We represent the function as $G: (R^z,\theta) \rightarrow R_{\{0,1\}}^{d_1}\times \cdots\times  R_{\{0,1\}}^{d_F}$ where $F$ is the number of categorical random variables and $d_i$ is the number categories for $i^{\mbox{th}}$ categorical random variable.
However, due to discrete nature of the output variables, we can not use re-parameterization trick to sample from $G(z,\theta)$ \cite{vae}. Thus, we resort to Gumbel-softmax trick to obtain  sample from the distribution while allowing the flow of gradient through the neural network \cite{gumbel}. We represent bid request as  $\bx = [\bx^1\oplus \cdots\oplus \bx^{F}]$ where $\bx^i$ is the one hot encoding of $i^{\mbox{th}}$ categorical random variable with probability distribution $\pi^i$. Using Gumbel-Max trick, we can sample from the distribution $\pi^i$ as $ \bx^i = \mbox{onehot}(\mbox{arg}\max_j [g_j +\log \pi^i_j])    $ where $g_j$ is sample from standard Gumbel distribution. As the argmax operator is not differentiable, we use soft-max to sample from the distribution $\pi^i$ while allowing to compute gradient. Thus,
$\bx^i_j \simeq \frac{ \exp((g_j +\log \pi^i_j)/\tau)                }{\sum_k \exp((g_k +\log \pi^i_k)/\tau)}    $
where $\tau$ is the temperature parameter with $\tau\rightarrow 0$ representing argmax. During training we keep $\tau>0$ to allow flow of gradient similar to \cite{gumbel}.

\paragraph{Market Action Model.}  Maximization of the $2^{\mbox{nd}}$ term in Eq. \ref{eq:main}, $\ell_2= \BE_{\ba^M\sim \bs^M|\bx } \log \Pr(\ba^M|\bx)$, denotes maximizing log-likelihood of market action given a bid request. Note, the market actions are a tuple of market price and the reward value $\ba^M = (\bw_i, \br_i)$. We solve when the utility $\br_i$ is $\mbox{click}_i$. The market price and a click from the user (when ad is shown) are independent of each other; thus we can write $\ell_2 =  \BE\log (\Pr(\bw|\bx))    +\BE\log (\Pr(\mbox{click}|\bx)).$ Note when $\br_i=\mbox{imp}_i$, we need to only solve  $\ell_2=  \BE\log (\Pr(\bw|\bx))$ as $\mbox{imp}_i$ is a deterministic function of market price $\bw_i$ and bid value $\ba_i$.

For the first term $\BE\log \Pr(\bw|\bx)$, we know the market price when the advertiser wins the auction ($\CW$) but only have lower bound when the advertiser loses ($\CL$). Thus we use fully parametric censored regression to estimate the market price distribution parameterized as $\CN(f_1(\bx), \exp(f_2(\bx))^2)$ where $f_1,f_2$ are deep neural networks \cite{cr,mdn-cr}. The objective is to maximize:
\begin{eqnarray*}
 \sum\log \Pr(\bw|\bx) = \sum_{\CL} \log \Pr(\bw\geq \mbox{bid}|\bx) + \sum_{\CW} \log \Pr(\bw|\bx) 
\end{eqnarray*}
The second term, $\BE\log (\Pr(\mbox{click}|\bx))$, is always observed when the user sees the ad impression. Thus, we maximize the binary classification problem,
$\BE\log (\Pr(\mbox{click}|\bx))$, using logistic regression on observed samples.

\subsection{Batch Policy Learning and Evaluation Framework}
The maximum entropy market model equips us with samples from the market state model $\bq_{\theta}(\bx)$ and market action model $\bp_w(\bw;\bx)$ $\bp_r(\br;\bw,\bx,\ba) $. With any starting advertiser state $\bs_0^A=[\bb_0\oplus\bt_0]$, the agent can simulate the next advertiser state $\bs_1^A$ by sampling  the market state $\bx_0\sim \bq_{\theta}$ and the market price $\bw_0$ from the action model using $1^{\mbox{st}}$ conditional independence in  Eq.~\ref{eq:1st}. The current market state $\bx_i$ and  price $\bw_i$ allow updating the next advertiser state $\bs_{i+1}^A$ while market state distribution $\bq_{\theta}$ allows sampling of the next market state $\bx_{i+1}$. The simulated training environment enables us to run any model-free RL algorithms to learn the optimal bidding strategy. Further, as the environment allows us to explore, we can start with any budget and time constraints for the advertiser to improve generalization beyond states observed in past experiences. In Algorithm ~\ref{alg:learn},  we outline the generic approach for training RL algorithms using the market state and the action model. Moreover using the market simulated (testing) environment, we can evaluate any model under any budget and time constraints. Similar to the training framework, the advertiser can start with any arbitrary state while allowing to evaluate any bidding strategy using the market model.  We outline the evaluation framework in Algorithm ~\ref{alg:test}.

\begin{algorithm}
   \algsetup{linenosize=\small}
  \small
    \caption{Generic Approach for Optimal Bidding}
    \label{alg:learn}
    \begin{algorithmic}[1]
        \STATE{\textbf{Input} Market state model $\bx\sim \bq^{\mbox{train}}_{\theta}(\bx)$, market action model $\bw \sim \bp^{\mbox{train}}_w(\bw;\bx)$, $\br\sim \bp^{\mbox{train}}_r(\br;\bw,\bx,\ba)$ }
        \STATE{\textbf{Input} Advertiser state  $\bs^D=[\bb\oplus \bt]$ }
        \STATE{\textbf{Input} RL Agent/Strategy with a random Policy}
        \WHILE{agent does not converge}
        \STATE{Sample market state $\bx\sim \bq^{\mbox{train}}_{\theta}(\bx)$, market price $\bw\sim \bp^{\mbox{train}}_w(\bw;\bx)$ }
        \STATE{Agent  bids $\ba$ for observed state $\bs^O= [\bx\oplus \bs^D]$ }
        \STATE{Agent observes reward $\br \sim \bp^{\mbox{train}}_r(\br; \bw,\bx, \ba)$}
        \STATE{Update Advertiser state ($[\bb\oplus\bt]$)}, Policy
        \ENDWHILE
        \RETURN Policy
    \end{algorithmic}
\end{algorithm}
\begin{algorithm}
\algsetup{linenosize=\small}
 \small
    \caption{Evaluation Framework}
    \label{alg:test}
    \begin{algorithmic}[1]
        \STATE{\textbf{Input} Market state model $\bx\sim \bq^{\mbox{test}}_{\theta}(\bx)$, market action model $\bw \sim \bp^{\mbox{test}}_w(\bw;\bx)$, $\br\sim \bp^{\mbox{test}}_r(\br;\bw,\bx,\ba)$ }
        \STATE{\textbf{Input} Advertiser state  $\bs^D=[\bb\oplus \bt]$ }
        \STATE{\textbf{Input} Agent/Strategy to evaluate}
        \STATE{Initialize Total Reward to $0$}
        \WHILE{episode did not end}
        \STATE{Sample market state $\bx\sim \bq^{\mbox{test}}_{\theta}(\bx)$, market price $\bw\sim \bp^{\mbox{test}}_w(\bw;\bx)$ }
        \STATE{Agent  bids $\ba$ for  observed state $\bs^O= [\bx\oplus \bs^D]$ }
        \STATE{Agent observes reward $\br \sim \bp^{\mbox{test}}_r(\br; \bw,\bx, \ba)$}
        \STATE{Update Advertiser state ($[\bb\oplus\bt]$)}, Total Reward
        \ENDWHILE
        \RETURN Total Reward
    \end{algorithmic}
\end{algorithm}
We use the Dueling Double Deep Q-network (DDQN) to train our agent \cite{dueling, ddqn} where we use a current $\psi$ network to train, a target $\psi^{-}$ network to compute the target value function and  a memory replay buffer $M$ to sample from old experience. We optimize:
\begin{equation}
\label{eq:policy}
\BEE_{(s,a,r,s')\sim M}  [(r+\gamma  Q_{\psi^-}(s',\argmax_{a'} Q_{\psi}(s',a')             )-Q_{\psi}(s,a  ))^2 ]
\end{equation}
Q function in DDQN is estimated using two separate branch of the neural network as $Q(s,a)= V(s)+ (A(s,a)- \frac{1}{|A|}\sum_a A(s,a)            )  $
where the advantage function is defined as, $A^{\pi} (s,a) = Q^{\pi}(s,a) -V^{\pi}(s)$. 
We quantize the one-dimensional continuous bidding values (actions) and use a discrete action space RL algorithm,  DDQN  for improved stability reason on long trajectories \cite{human,dueling}.
\vspace{-0.2cm}
\begin{remark} Computational latency is important in real-time bidding system. The final policy $\pi^{\ast}$ takes bid request $\bx$, budget left $\bb$, and, time left $\bt$ as input and computes the optimal bid price $\ba$ using neural network $\psi$ in Eq.~\ref{eq:policy}. The latency depends on the neural architecture; we use a simple three hidden layer neural network. One can easily augment the policy with a memory state to encode the past using a recurrent neural network. However, in such cases, the bidding platform needs to store memory state for each advertiser in addition to budget and time constraints. 
\end{remark}

%% file: tikz_summary.tex
\begin{figure*}
	\begin{center}
	\pgfmathsetmacro{\circ}{.35}
	     \scalebox{.9}{
		\begin{tikzpicture}[xscale=1, yscale=1]
		\draw [rounded corners][dashed] (-9.5,-1.5) rectangle (-2.5,-1.3+\circ+1.5);		
		\node at (-7.5, -1.25) {\textbf{\tiny State  $\mathbf{\bs_i = [\bs_i^M\oplus \bs_i^A]}$}};
		\draw [fill=yellow!0] (-9,-1.3+\circ) circle [radius=\circ];;
	    \node at (-9,-1.3+\circ) { {\tiny $\mathbf{\bs_i^A}$}};
	    
	    \draw [fill=yellow!0] (-6,-1.3+\circ) circle [radius=\circ];;
	    \node at (-6,-1.3+\circ) { \textbf{\tiny $\mathbf{\bs_{i+1}^A}$}};
		
		\draw [fill=yellow!0] (-3,-1.3+\circ) circle [radius=\circ];;
	    \node at (-3,-1.3+\circ) { \textbf{\tiny $\mathbf{\bs_{i+2}^A}$}};
	    
	    \draw [->] (-9+\circ,-1.3+\circ) to (-6-\circ,-1.3+\circ);
	    \draw [->] (-6+\circ,-1.3+\circ) to (-3-\circ,-1.3+\circ);
	    
	    \draw [->] (-9+\circ,-1.3+1+\circ) to (-6-1*\circ,-1.3+1.5*\circ);
	    \draw [->] (-6+\circ,-1.3+1+\circ) to (-3-\circ,-1.3+1.5*\circ);
	    
	    \draw [->] (-9+0.5*\circ,-1.3+2.25+0.5*\circ) to (-6,-1.3+2*\circ);
	    \draw [->] (-6+0.5*\circ,-1.3+2.25+0.5*\circ) to (-3,-1.3+2*\circ);
		\draw [fill=orange!0] (-9,-1.3+\circ+1) circle [radius=\circ];;
	    \node at (-9,-1.3+\circ+1) { \textbf{\tiny $\mathbf{\bs_i^M}$}};
	    
	    \draw [fill=orange!0] (-6,-1.3+\circ+1) circle [radius=\circ];;
	    \node at (-6,-1.3+\circ+1) { \textbf{\tiny $\mathbf{\bs_{i+1}^M}$}};
		
		\draw [fill=orange!0] (-3,-1.3+\circ+1) circle [radius=\circ];;
	    \node at (-3,-1.3+\circ+1) { \textbf{\tiny $\mathbf{\bs_{i+2}^M}$}};
	    
		\draw [fill=blue!0] (-9,-1.3+\circ+2.25) circle [radius=\circ];;
	    \node at (-9,-1.3+\circ+2.25) { \textbf{\tiny $\mathbf{\ba_i}$}};
	    
	    \draw [fill=blue!0] (-6,-1.3+\circ+2.25) circle [radius=\circ];;
	    \node at (-6,-1.3+\circ+2.25) { \textbf{\tiny $\mathbf{\ba_{i+1}}$}};
		
		\draw [fill=blue!0] (-3,-1.3+\circ+2.25) circle [radius=\circ];;
	    \node at (-3,-1.3+\circ+2.25) { \textbf{\tiny $\mathbf{\ba_{i+2}}$}};
	    
	    \draw [->] (-9+\circ,-1.3+\circ) to [out=0,in=0] (-9+\circ,-1.3+\circ+2.25);
	    \draw [->] (-6+\circ,-1.3+\circ) to [out=0,in=0] (-6+\circ,-1.3+\circ+2.25);
	    \draw [->] (-3+\circ,-1.3+\circ) to [out=0,in=0] (-3+\circ,-1.3+\circ+2.25);
	    
	    \draw [->] (-9,-1.3+2*\circ+1) to (-9,-1.3+2.25);
	    \draw [->] (-6,-1.3+2*\circ+1) to (-6,-1.3+2.25);
	    \draw [->] (-3,-1.3+2*\circ+1) to (-3,-1.3+2.25);
		\draw [fill=cyan!0] (-9,-1.3+\circ+3.75) circle [radius=\circ];;
	    \node at (-9,-1.3+\circ+3.75) { \textbf{\tiny $\mathbf{\br_i}$}};
	    
	    \draw [fill=cyan!0] (-6,-1.3+\circ+3.75) circle [radius=\circ];;
	    \node at (-6,-1.3+\circ+3.75) { \textbf{\tiny $\mathbf{\br_{i+1}}$}};
		
		\draw [fill=cyan!0] (-3,-1.3+\circ+3.75) circle [radius=\circ];;
	    \node at (-3,-1.3+\circ+3.75) { \textbf{\tiny $\mathbf{\br_{i+2}}$}};
		
		\draw [->] (-9,-1.3+2*\circ+2.25) to (-9,-1.3+3.75);
	    \draw [->] (-6,-1.3+2*\circ+2.25) to (-6,-1.3+3.75);
	    \draw [->] (-3,-1.3+2*\circ+2.25) to (-3,-1.3+3.75);
	    
	    \draw [->] (-9+\circ,-1.3+1+\circ) to [out=0,in=0] (-9+\circ,-1.3+\circ+3.75);
	    \draw [->] (-6+\circ,-1.3+1+\circ) to [out=0,in=0] (-6+\circ,-1.3+\circ+3.75);
	    \draw [->] (-3+\circ,-1.3+1+\circ) to [out=0,in=0] (-3+\circ,-1.3+\circ+3.75);
	    
		\draw [rounded corners][fill =cyan!10] (-1,2) rectangle (1,3);		
		\node at (0,2.5) {{\tiny Train Batch Data}};
				
		\draw [rounded corners][fill=red!10] (-1,.5) rectangle (1,1.5);
		\node at (0,1.) [align = center]{\tiny Maximum Entropy\\ \tiny Market Model};
		
		\draw [rounded corners][fill= blue!10] (1.5,.5) rectangle (3,1.5);
		\node at (2.25, 1) [align = center]{\tiny Advertise\\ \tiny State};
\draw [dashed][->] (1,1) to (1.5, 1);
		\draw [rounded corners][fill=yellow!10] (1.5,.5+1.5) rectangle (3,1.5+1.5);
		\node at (2.25, 1+1.5) [align = center]{\tiny Advertiser\\ \tiny  Initial State};
		
		\draw [rounded corners][] (-1.25,0) rectangle (3.25,1.75);		
		\node at (1,0.25) {{\tiny Train Environment}};
		
		\draw [rounded corners][fill=orange!10] (0,-1.5) rectangle (2, -1);
		\node at (1, -1.25) {{\tiny Train Agent}};
		
		\draw [->] (0,2) to (0, 1.5);
		\draw [->] (2.25,2) to (2.25, 1.5);
		\draw [->] (.25,0) to (.25, -1);
		\node at (.1, -.5) {{\tiny $s_t$}};
		\draw [<-] (1,0) to (1, -1);
		\node at (.9, -.5) {{\tiny $a_t$}};
		\draw [->] (1.75,0) to (1.75, -1);
		\node at (1.65, -.5) {{\tiny $r_t$}};

\draw [rounded corners][fill =cyan!10] (4,2) rectangle (6,3);		
\node at (5,2.5) {{\tiny Test Batch Data}};

\draw [rounded corners][fill=red!10] (4,.5) rectangle (6,1.5);
\node at (5,1.) [align = center]{\tiny Maximum Entropy\\ \tiny Market Model};

\draw [rounded corners][fill= blue!10] (6.5,.5) rectangle (8,1.5);
\node at (7.25, 1) [align = center]{\tiny Advertise\\ \tiny State};

\draw [dashed][->] (6,1) to (6.5, 1);
\draw [rounded corners] (3.75,0) rectangle (8.25,1.75);		
\node at (6,0.25) {{\tiny Test Environment}};

\draw [rounded corners][fill=yellow!10] (1.5+5,.5+1.5) rectangle (3+5,1.5+1.5);
\node at (2.25+5, 1+1.5) [align = center]{\tiny Advertiser\\ \tiny  Initial State};

		\draw [->] (0+5,2) to (0+5, 1.5);
			\draw [->] (2.25+5,2) to (2.25+5, 1.5);
\draw [rounded corners][fill=orange!10] (0+5,-1.5) rectangle (2+5, -1);
\node at (1+5, -1.25) {{\tiny Test Agent}};
\draw [->] (.25+5,0) to (.25+5, -1);
\node at (.1+5, -.5) {{\tiny $s_t$}};
\draw [<-] (1+5,0) to (1+5, -1);
\node at (.9+5, -.5) {{\tiny $a_t$}};
\draw [->] (1.75+5,0) to (1.75+5, -1);
\node at (1.65+5, -.5) {{\tiny $r_t$}};

\draw [dashed] (2,-1.25) to (5, -1.25);

		
		

		\end{tikzpicture} }
	\end{center}
\caption{a) Graphical model of the Bidding System (left). b) Training and Evaluation Framework (right). Agent optimized using train environment is tested on test  environment.}
\label{fig:summary}
\end{figure*}
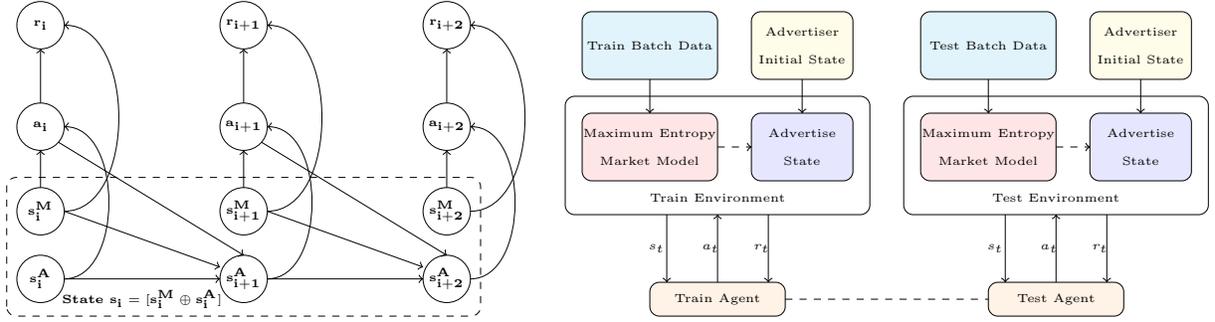

%% file: exp.tex
\section{Experimental Results}
\subsection{Experimental Setup}
\begin{figure*}
    \subfloat[Advertiser 1458]{\includegraphics[width = 0.2\linewidth]{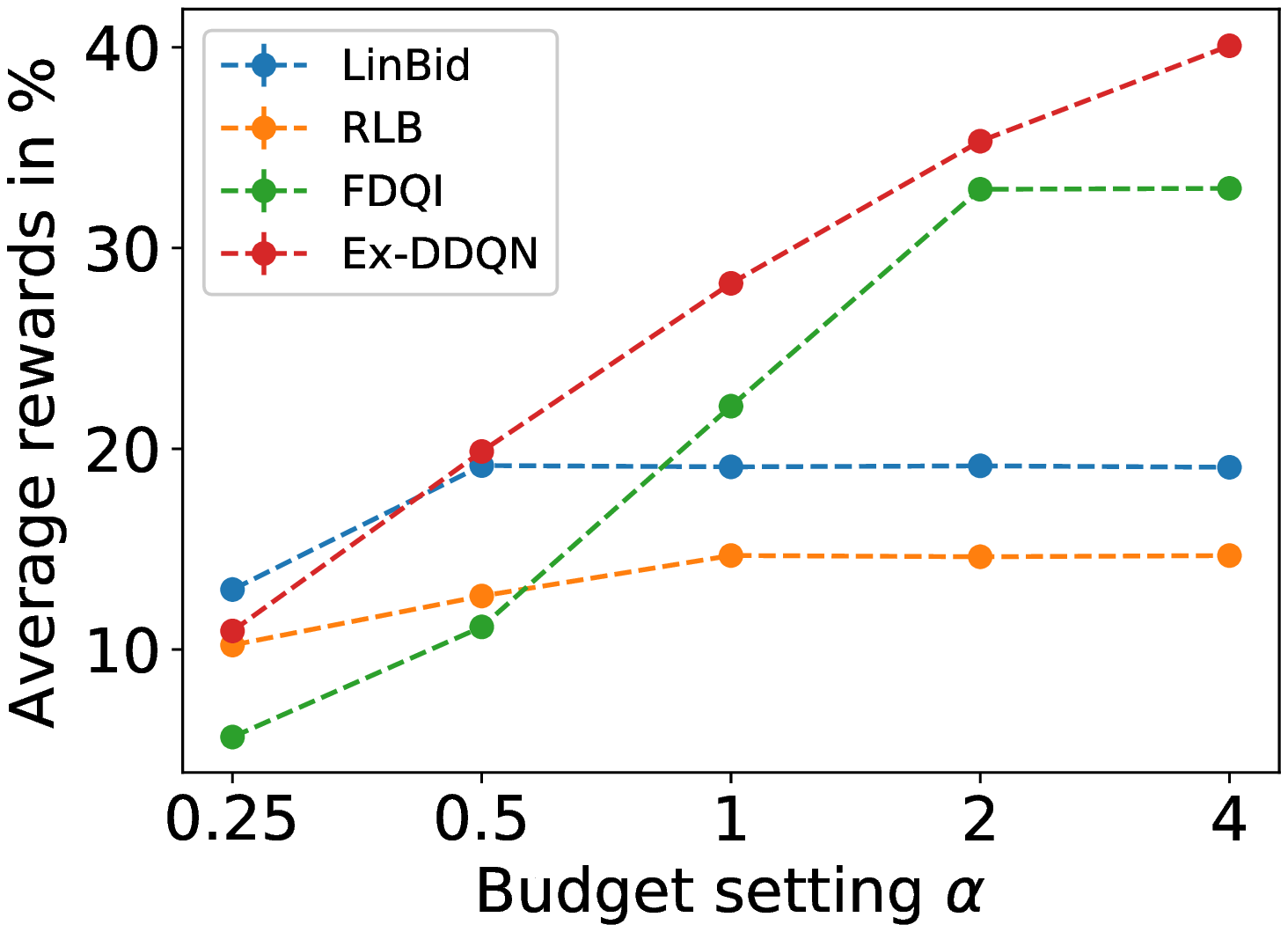}} 
    \subfloat[Advertiser 2259]{\includegraphics[width = 0.2\linewidth]{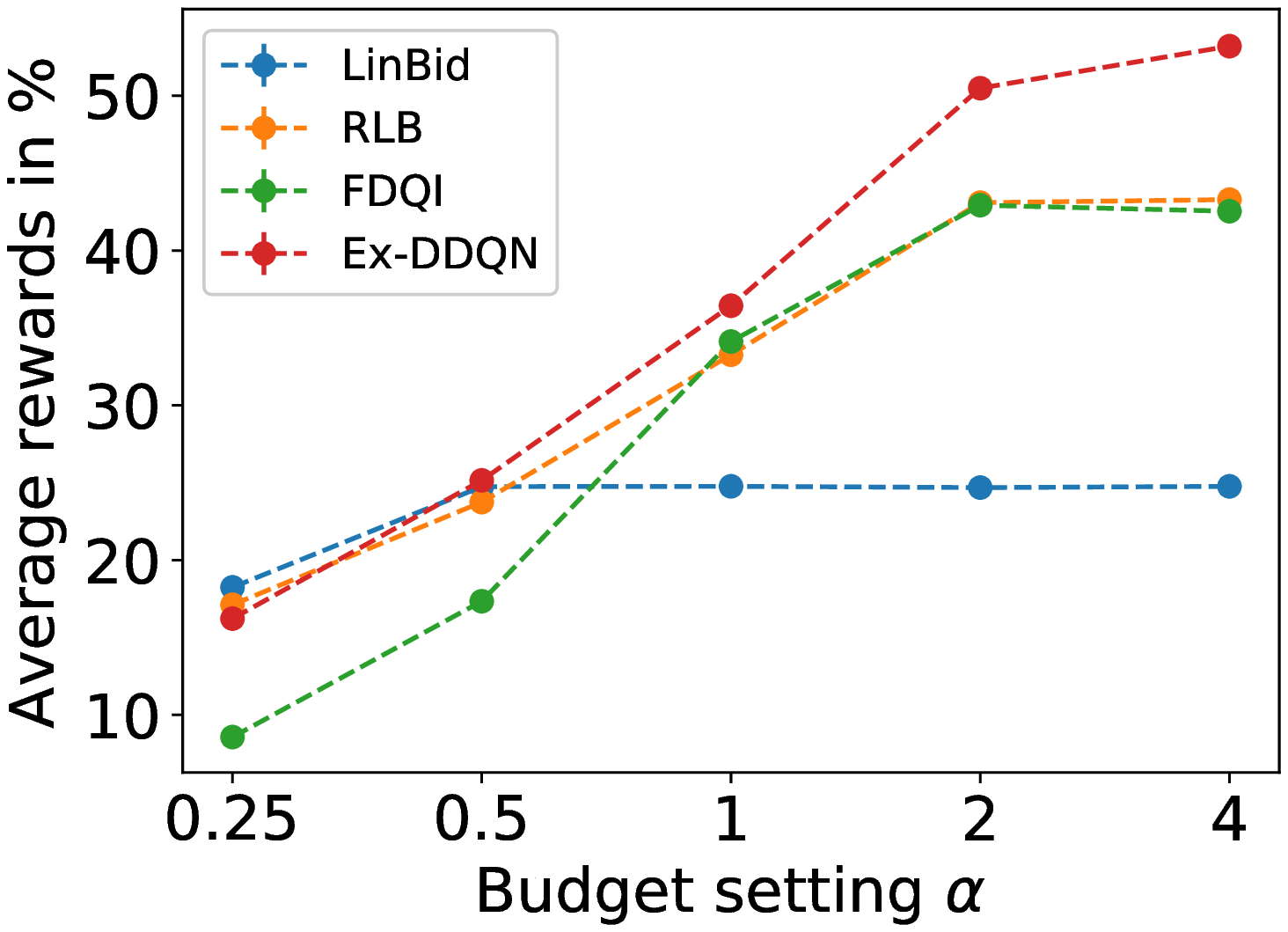}}
    \subfloat[Advertiser 2261]{\includegraphics[width = 0.2\linewidth]{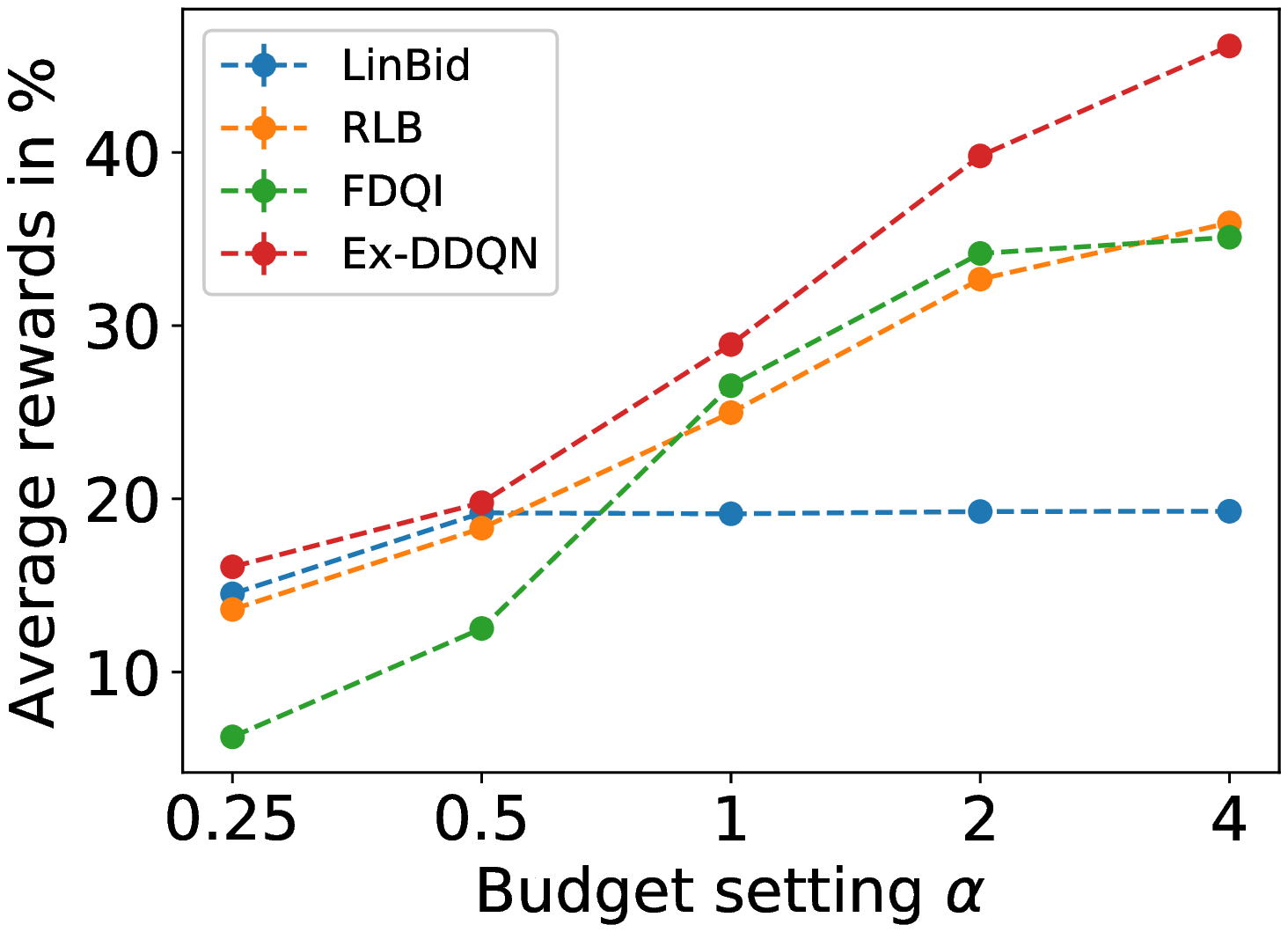}}
    \subfloat[Advertiser 2821]{\includegraphics[width = 0.2\linewidth]{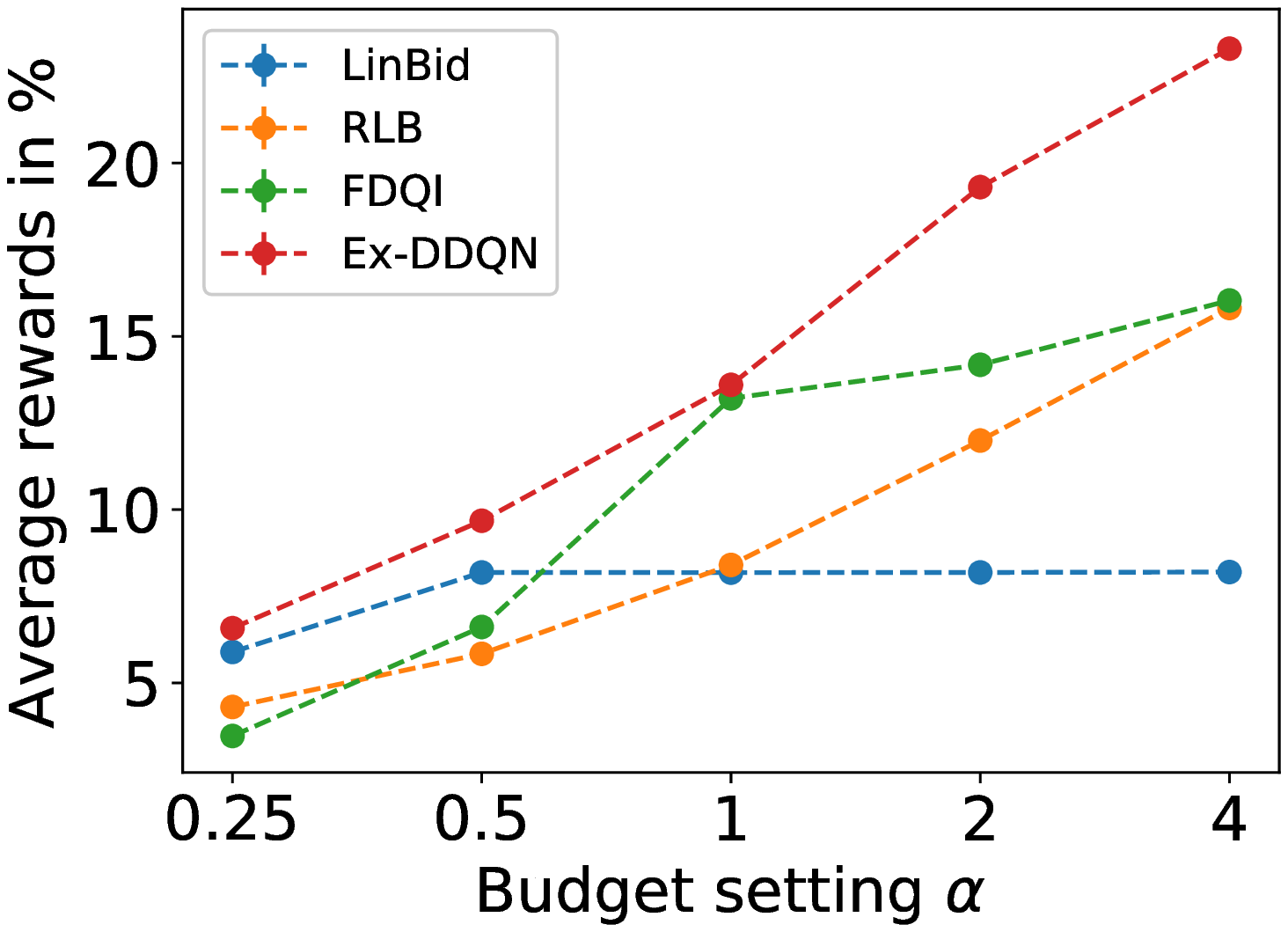}} 
    \subfloat[Advertiser 2997]{\includegraphics[width = 0.2\linewidth]{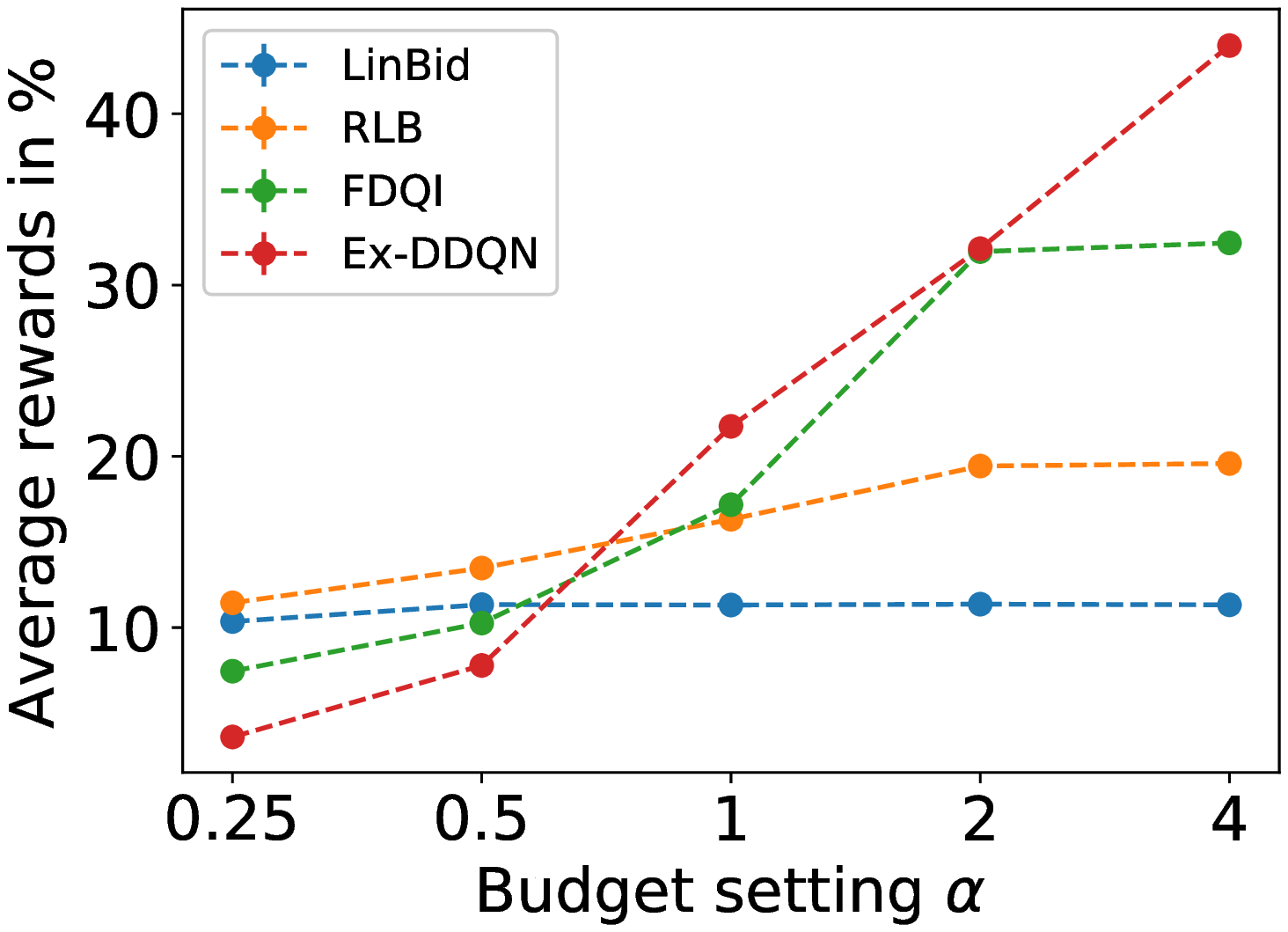}}\\ 
    \subfloat[Advertiser 3358]{\includegraphics[width = 0.2\linewidth]{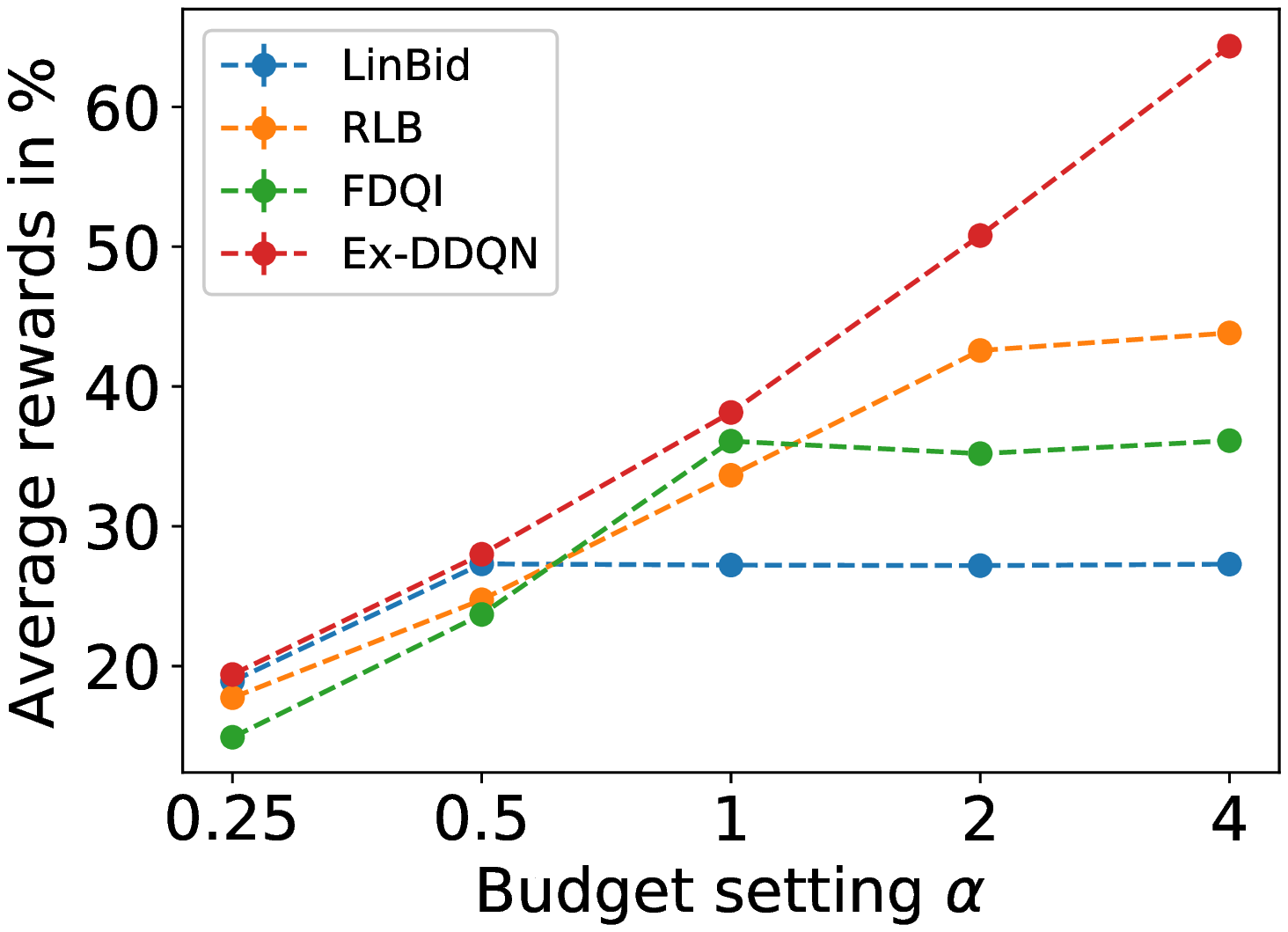}} 
    \subfloat[Advertiser 3386]{\includegraphics[width = 0.2\linewidth]{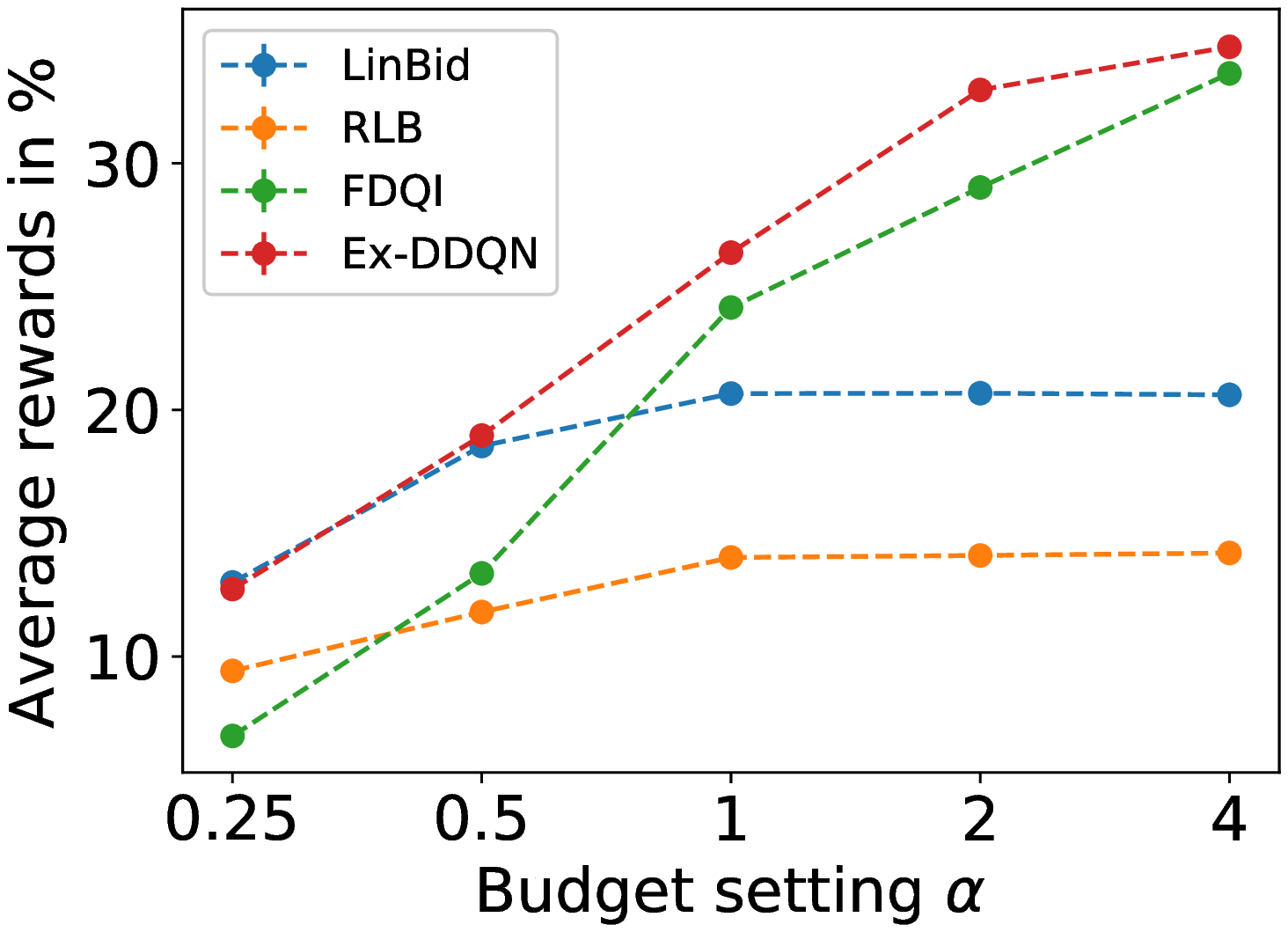}} 
    \subfloat[Advertiser 3427]{\includegraphics[width = 0.2\linewidth]{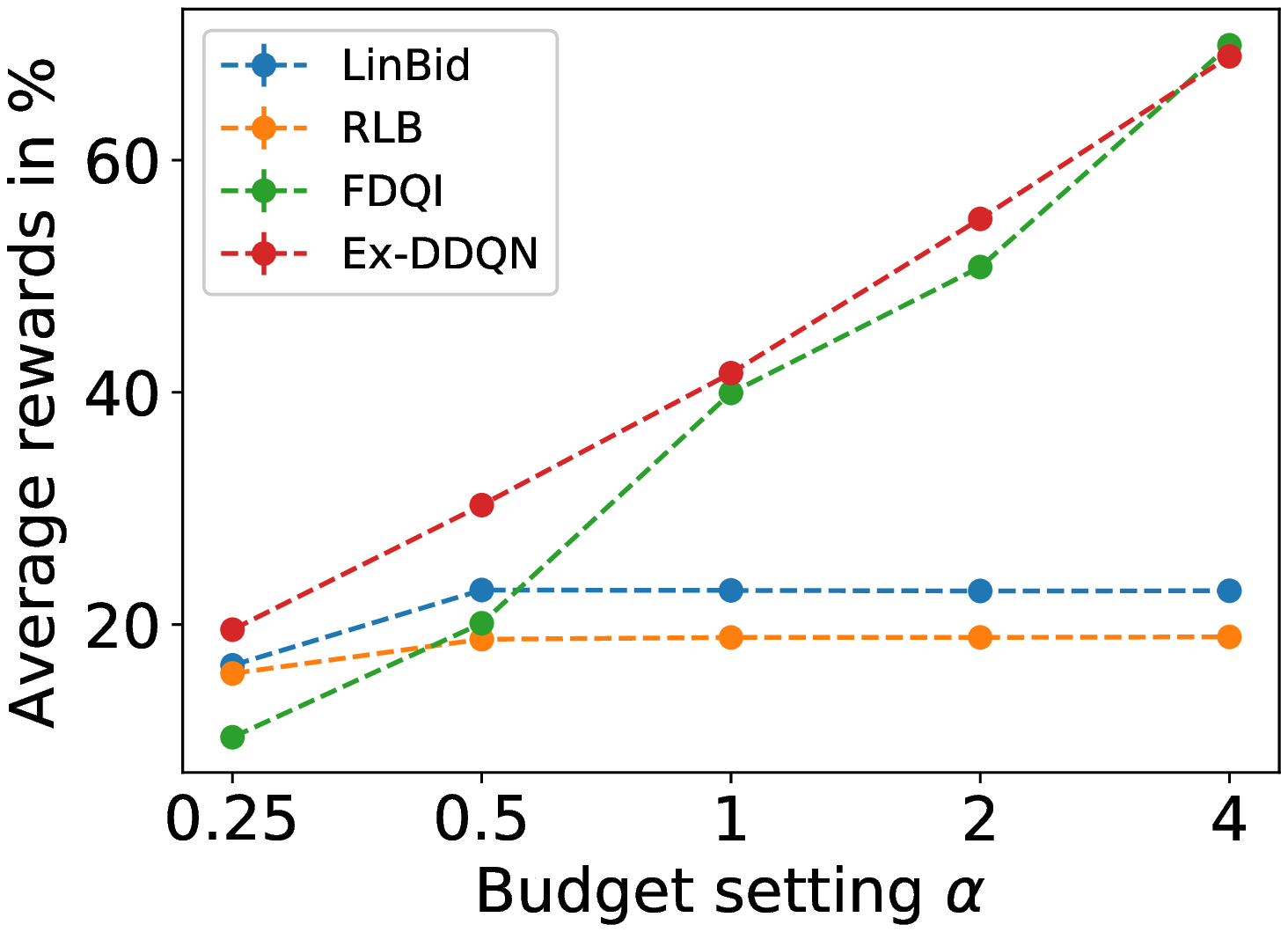}}
 \subfloat[Advertiser 3476]{\includegraphics[width = 0.2\linewidth]{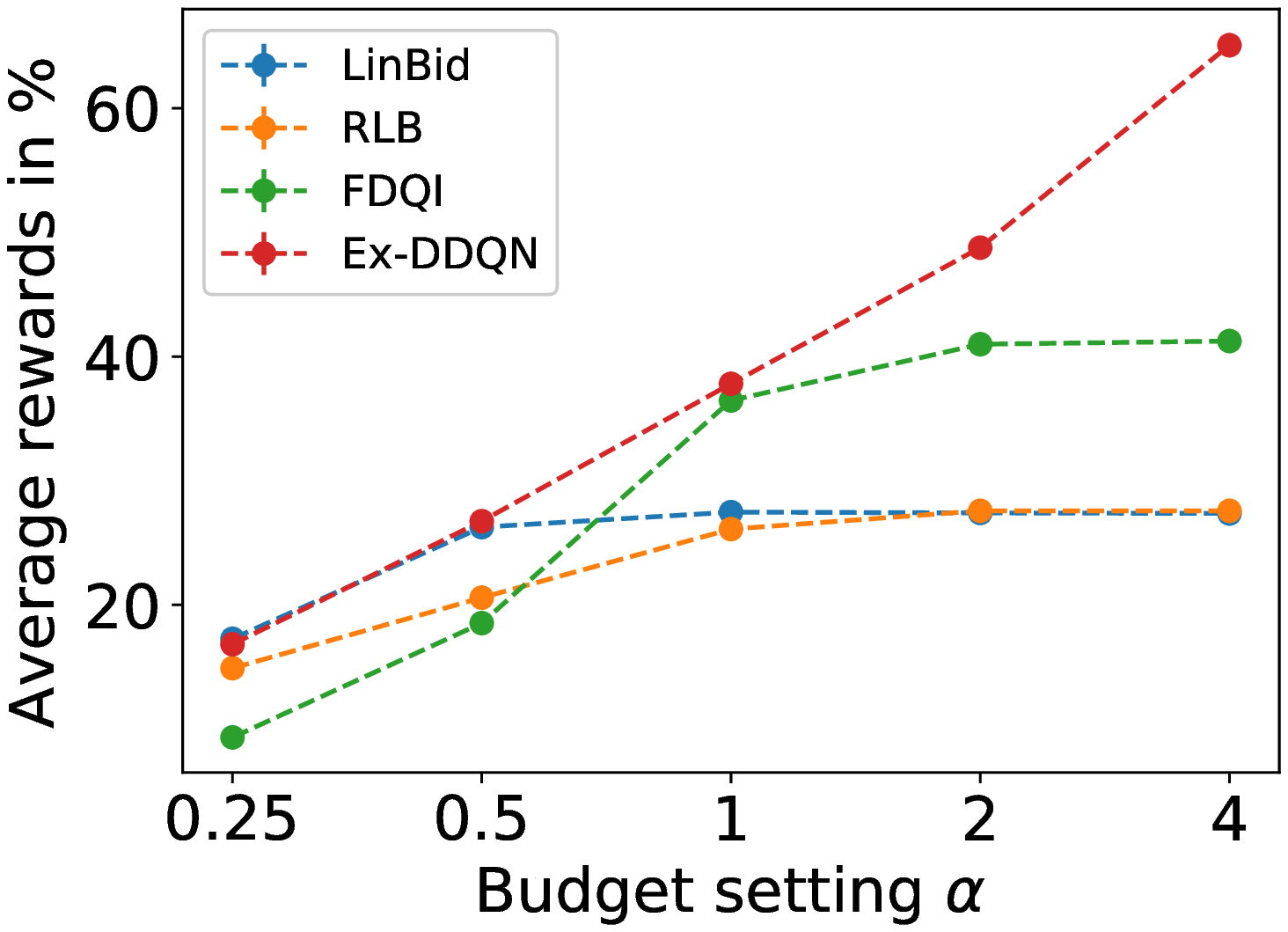}}             \subfloat[Adobe]{\includegraphics[width = 0.2\linewidth]{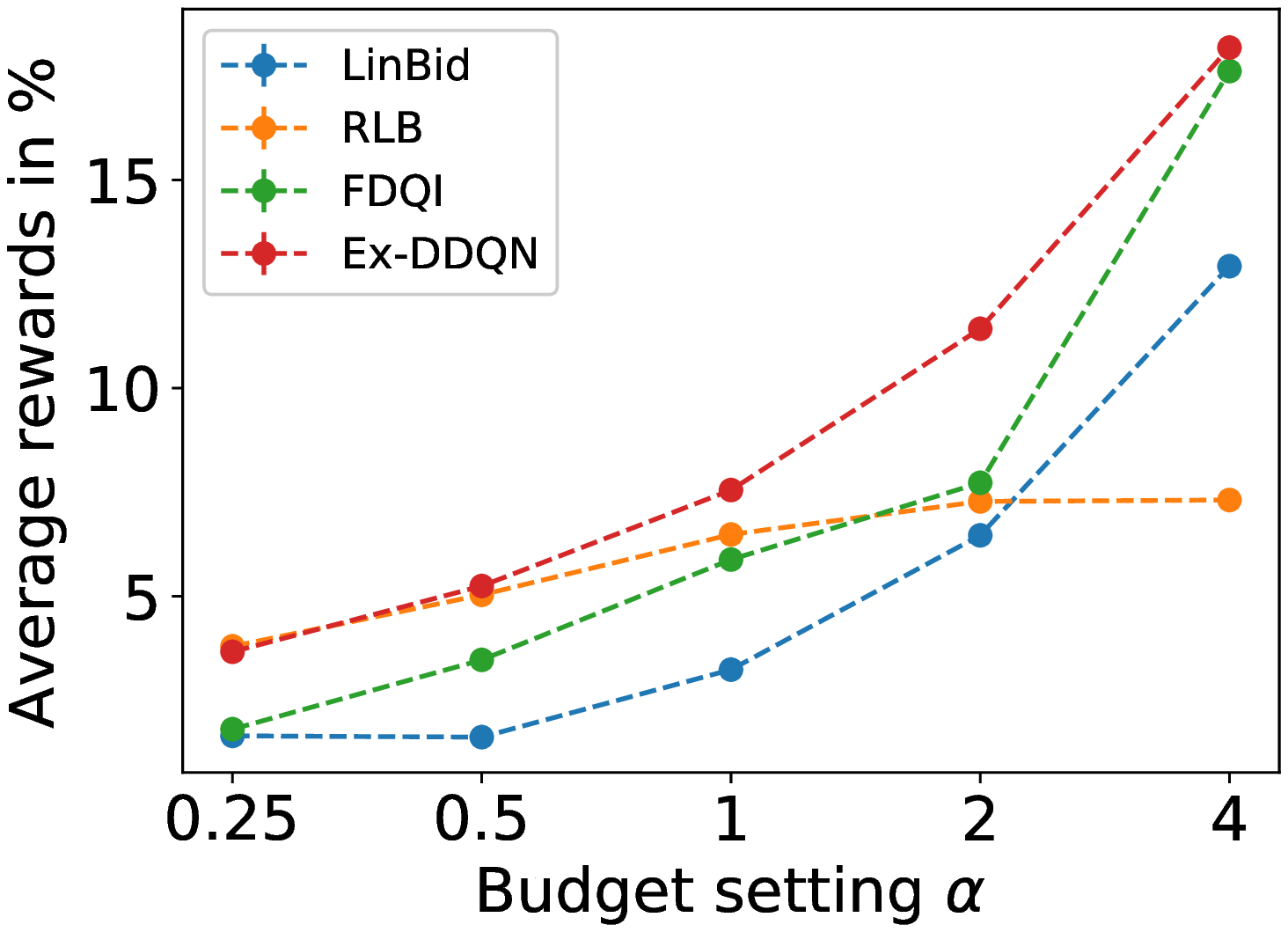}}
    \caption{Average Rewards (Impressions) under varied budget settings on iPinYou Advertisers and Adobe Dataset. Standard deviation (std) is on the order of $10^{-1}$.
    \label{fig:result1}}
\end{figure*}

 \begin{table}[t]\centering
    \scalebox{0.78}{
        \begin{tabular}{c  ccccccc}
        \hline
            {\footnotesize  Adv} & {\footnotesize $n\times 10^5$} & {\footnotesize  d} & {\footnotesize  $\mbox{imp}_{tr}$} & {\footnotesize $\mbox{imp}_{te}$} & {\footnotesize $\cpmtrain$} & {\footnotesize $\cpmtest$} & {\footnotesize $\mbox{KL}(\hat{P}_{tr}||\hat{P}_{te})$} \\ \hline
1458 & $147$ & $2140$ & $0.172$ & $0.302$ & $11.9$ & $20.7$ & $0.057$ \\
2259 & $14$ & $1097$ & $0.65$ & $0.412$ & $56.8$ & $40.6$ & $0.141$ \\
2261 & $12$ & $992$ & $0.602$ & $0.349$ & $52.1$ & $30.0$ & $0.179$ \\
2821 & $29$ & $1919$ & $0.548$ & $0.166$ & $47.5$ & $16.5$ & $0.46$ \\
2997 & $7$ & $428$ & $0.359$ & $0.301$ & $21.4$ & $19.0$ & $0.012$ \\
3358 & $37$ & $1875$ & $0.604$ & $0.405$ & $55.6$ & $39.4$ & $0.171$ \\
3386 & $140$ & $2054$ & $0.168$ & $0.276$ & $12.5$ & $22.0$ & $0.049$ \\
3427 & $140$ & $3970$ & $0.126$ & $0.442$ & $10.1$ & $35.9$ & $0.242$ \\
3476 & $67$ & $1662$ & $0.257$ & $0.419$ & $20.7$ & $31.9$ & $0.078$ \\ \hline
        \end{tabular}
    }
    \caption{Dataset statistics for iPinYou Advertisers.}
    \label{tab:dataset}
\end{table}

\paragraph{Datasets.} We use two real-world datasets to conduct our experiments. Publicly available {\bf iPinYou} dataset contains bidding data from $9$ advertisers over multiple days \cite{ipinyou}. Bid requests contain multiple categorical features such as city, domain, etc. 
Table ~\ref{tab:dataset} lists total number of bid requests ($n\times 10^5$), categorical features ($d$), average train and test impression ($\mbox{imp}_{tr},\mbox{imp}_{te}$), training and testing cost  per thousands bid requests ($\cpmtrain, \cpmtest$), and, the KL divergence between train and test empirical market price distribution (where a higher number implies market price distribution in train and test datasets are more different). 
 We learn the optimal bidding strategy from the perspective of each advertiser on the iPinYou dataset. For experimental purposes, we also collect a fraction of bid requests from a week's data from \textbf{Adobe}, one of the leading DSP. The bid requests contain categorical features similar to iPinYou.  We learn an aggregate optimal bidding strategy for the Adobe DSP without splitting the advertisers. The total number of bid requests and categorical features are $2,323,226$ and $1282$ respectively.
 
 \paragraph{Baseline Methods} We compare the following methods using the same testing environment with varied budget constraints.

{\bf LinBid \cite{linbid}}  is the linear bidding strategy with $bid = b_0 \theta(\bx)$ where $b_0$ is the base parameter tuned on the training dataset to maximize total rewards. 

{\bf RLB-Segmentation \cite{rlb}} is the state-of-the-art model that solves the Bellman equation using model approximation. Further, RLB uses coarse to fine segmentation to handle large sequences. 

{\bf FDQI \cite{fdqn}} is a generic method for batch RL using neural fitted deep Q-iteration. Although early works in real-time bidding systems do not use FDQI, we add this general batch RL method as a baseline.

{\bf Ex-DDQN} is our proposed approach based on the market environment for exploration. We learn the training market environment using the maximum entropy principle. We train model-free dueling Q-network with varied budget constraints using the learned training environment.

\paragraph{Training and Testing Environment.} For each advertiser in the iPinYou dataset, we use the $60\%,\ 15\%,\ 25\%$ of the days' samples as the training, validation and testing data respectively. For the Adobe dataset, instead of days, we randomly split the samples as the training, validation, and testing batch dataset with the same fraction. For evaluation, we learn the simulated market environment using only the test datasets. In particular, we learn market state model, impression and click models from test datasets (we use validation datasets to do early stopping). For training, all algorithms use training datasets. For exploration during training, Ex-DDQN uses the simulated market environment learned using training datasets.

 We learn the market state model using the Wasserstein Generative Adversarial Network with gradient penalty (WGAN-GP) \cite{wgan-gp}. To learn the impression model, we maximize fully parametric censored regression \cite{mdn-cr}. We learn the click model using logistic regression. The impression and click model allows us to sample market price and user click behavior respectively. To generate a bid request, we sample from a standard normal distribution as the input to the generator of the WGAN-GP. We learn these three models to simulate the market environment from the train and test datasets (for training and evaluation purposes respectively).  
 Note, \emph{in no cases}, learning training and the testing environment share any samples.
 Network architecture, featurization, hyper-parameters used for tuning can be found in the  long version of this paper. 
\paragraph{Evaluation Metric}
We consider the number of \textit{total impressions}, over the trajectories, as the utility of interest. Thus, given a budget and time constraint, the goal is to maximize the number of impressions for the advertiser. Note, previous research often considers \textit{click} as the utility removing all non-impressed bid requests \cite{rlb}. However, for a click to occur, the advertiser needs to win the auction first. Our framework allows both click and impression as the utility measure. However, as clicks are sparse ($\sim0.1\%$), we choose to tabulate results with impressions as the utility measure to reduce variance in the estimation. 
Further, we consider a more realistic large scale evaluation where we optimize bidding strategy for a sequence of $T_0=100,000$ bid requests, which correspond to real-world $10$ minutes auction volume in case of iPinYou \cite{rlb,ipinyou}.

To evaluate generalization properties of any bidding strategy, we set budget as $B_0 = \alpha \times\cpmtest \times \frac{T_0}{10^3}$ where $\cpmtest$ is the average cost of the advertiser on the test dataset over $1000$ bid requests. We set $\alpha = \{\frac{1}{4}, \frac{1}{2}, 1, 2, 4\}$ as the budget constraint. Previous research only considers $\alpha =\{\frac{1}{32}, \frac{1}{16}, \frac{1}{8}, \frac{1}{4}, \frac{1}{2}\}$, and, set test datasets where only the advertisers win the impression. However, this setting is very unrealistic in real-world scenarios; besides the test distribution shift, previous works consider cases where the budget is strictly less than the spent \cite{rlb}. In our evaluation, we do not directly use the test dataset; instead, we learn a  test market sampler and use samples from the sampler to test all model fairly. For all budget settings, we repeat the experiment $10$ times to get the average and the standard deviation (std) numbers.
 \begin{table}[t]\centering
    \scalebox{0.8}{
        \begin{tabular}{c ccc}
        \hline
            Advertiser & Test Sample & Model Sample & Uniform Sample \\ \hline
            1458 & $0.138\pm 0.013 $ & $ 0.253\pm 0.028 $ & $ 2.622\pm 0.123 $\\
2259 & $0.142\pm 0.016 $ & $ 0.182\pm 0.021 $ & $ 2.668\pm 0.118 $\\
2261 & $0.141\pm 0.013 $ & $ 0.166\pm 0.017 $ & $ 2.408\pm 0.079 $\\
2821 & $0.143\pm 0.015 $ & $ 0.184\pm 0.026 $ & $ 2.857\pm 0.098 $\\
2997 & $0.145\pm 0.019 $ & $ 0.15\pm 0.019 $ & $ 4.57\pm 0.187 $\\
3358 & $0.141\pm 0.014 $ & $ 0.339\pm 0.049 $ & $ 2.097\pm 0.065 $\\
3386 & $0.141\pm 0.013 $ & $ 0.228\pm 0.039 $ & $ 2.732\pm 0.128 $\\
3427 & $0.144\pm 0.013 $ & $ 0.2\pm 0.022 $ & $ 2.158\pm 0.108 $\\
3476 & $0.143\pm 0.014 $ & $ 0.172\pm 0.015 $ & $ 1.99\pm 0.108 $\\
Adobe & $0.143\pm 0.013 $ & $ 0.156\pm 0.015 $ & $ 1.558\pm 0.063 $\\
\hline
        \end{tabular}
    }
    \caption{$\sqrt{n}\times \hat{\mbox{MMD}}$ Distance between Test samples and $\{\mbox{Test, Model, Uniform}\}$ samples. $n =200$.}
    \label{tab:mmd}
\end{table}
\subsection{Results}
In Figure ~\ref{fig:result1}, we plot the percentage of impressions won over the trajectories ($100,000$ timesteps) for all advertisers on the iPinYou dataset as well as on the Adobe dataset with different budget constraints. 

On the lowest budget settings ($\mbox{budget} =\alpha \times\cpmtest \times \frac{T_0}{10^3}$) with $\alpha=\frac{1}{4}$, all batch RL methods, RLB, LinBid, FDQI have experiences from the historical interactions; thus we expect relatively better performance from these algorithms. For all advertisers, except for advertiser 2997, Ex-DDQN consistently performs similar or better than RLB/LinBid methods while significantly improving on the FDQI method. Performances of RLB and Linbid are mixed with cases one works better than the other and vice-versa. 
For budget setting $\alpha =\frac{1}{2}$, Ex-DDQN performs significantly better than all algorithms on nine of the ten cases. The average gain over the next best performance on the nine advertisers is more than $7\%$.
While FDQI does not perform well, the other two batch algorithms, LinBid, and, RLB perform relatively well with a few cases improving upon Ex-DDQN. The result suggests Ex-DDQN performs similar (and sometimes better) to the state-of-the-art method when budget settings are small ($\alpha\leq 1/2$).

For budget settings $\alpha\geq 1$, we compare the generalization performance of the batch RL algorithms. Interestingly in all cases, Ex-DDQN significantly (sometimes with margin $\geq 10\%-30\%)$ improves LinBid and RLB algorithms while FDQI follows Ex-DDQN performance.  The significant performance gain suggests agents trained to explore with varied advertiser states on the simulated training environment can potentially generalize on the new test environment (even where train and test datasets are significantly different, see Table ~\ref{tab:dataset}).

We further note that RLB performs relatively better when $\alpha\leq 1/2$ and KL divergence (listed in Table ~\ref{tab:dataset}) between empirical train and test market price is small. RLB being a model-based RL algorithm solves Bellman equation with empirical training market price; thus making them sensitive to model mismatch error. For example, on advertiser 2821 and 3427 (where KL divergences are particularly high), RLB performs significantly worse than even the LinBid algorithm. Both RLB and Linbid have saturating behavior when budget parameter $\alpha$ is more than $1$. The saturating behavior suggests that these batch models face difficulty extrapolating optimal behavior with higher budget constraints. 

\subsection{Market Model Evaluation}
We evaluate our testing environment to compare whether the generated model bid requests reflect the true samples from the test distribution. We sample $n=200$ bid requests from test batch dataset and sample another 200 bid requests each from test batch dataset, maximum entropy market model and uniformly randomly from the categorical features to compute empirical Maximum Mean Discrepancy (MMD) distance \cite{kernel}. In Table~\ref{tab:mmd}, we list the average empirical MMD distance and standard deviation from repeating the process 100 times. We use Gaussian kernel with $\sigma=1$. Empirical MMD distance is close to zero for samples between test batch datasets and samples from the market model while for a uniformly random sample, distance is an order of magnitude larger. This result validates our testing environment learned using the maximum entropy principle being close to the actual one in real-traffic. 

%% file: con.tex
\section{Discussion} In this paper, we tackle one of the key challenges in the real-time bidding system, learning the optimal bidding strategy from a batch dataset.  We propose a generic framework for evaluating models from batch datasets which were lacking in previous literature. Further, we propose a solution to learn the optimal strategy from the batch dataset that has the property to generalize to unseen state space in addition to competitive (sometimes better) performance on the known regime. Although we analyze only on RTB systems, learning from the batch dataset poses a significant challenge to any recommender system-based solutions. Without interacting with the user (or market) in real-traffic, evaluation/training becomes difficult. Potentially our framework can be applied in recommender systems where state and transition function decomposes similar to RTB systems; such extensions we leave for future work.

%% file: supplementary.tex
\section{Featurization} The public dataset iPinYou has bid, impression, click, and conversion logs. We did a join with bid logs and impression logs to compute the market price (in case of winning auctions). iPinYou data is grouped into two subsets: session 2 (dates from 2013-06-06 to 2013-06-12), and session 3 (2013-10-19 to 2013-10-27). We divide the whole dataset among the $9$ advertisers as done in previous research \cite{rlb}. We use the following fields: Timestamp, UserAgent, Region, City, AdExchange, Domain, AdSlotId, SlotWidth, SlotHeight, SlotVisibility, SlotFormat, Usertag. Every categorical feature (e.g City), is one-hot encoded, whereas every numerical feature (e.g Adheight) is categorized into bins and subsequently represented as one-hot encoded vectors.  This way, each bid request is represented as a large sparse vector
We converted Timestamp into two features namely the day of the week and hour of the day.
We converted User-Agent as a combination of oses (``windows", ``ios", ``mac", ``android", ``linux") and browsers (``chrome", ``sogou", ``maxthon",
    ``safari", ``firefox", ``theworld", ``opera", ``ie"). For each field, we use the features (field values such as City: 16) only which occurs 500 times in the dataset (with respect to the advertiser). 
For Adobe dataset, we have similar Categorical as well as Numerical features. We use the same procedure with a feature threshold of $1000$ to compute the sparse feature representation.

\begin{table}[t]
    \centering
    \scalebox{.7}{
        \begin{tabular}{ll llll}
            \hline 
            Adv     & Budget &  LinBid & RLB  & FDQI  & Ex-DDQN  \\ 
            \hline
\multirow{ 5}{*}{1458} & $\frac{1}{4}$  & $12.99\pm 0.09$  & $10.22\pm 0.07$  & $5.64\pm 0.04$  & $10.92\pm 0.11$ \\
& $\frac{1}{2}$  & $19.16\pm 0.14$  & $12.67\pm 0.07$  & $11.12\pm 0.09$  & $19.86\pm 0.1$ \\
& $1$  & $19.1\pm 0.12$  & $14.68\pm 0.11$  & $22.12\pm 0.08$  & $28.24\pm 0.08$ \\
& $2$  & $19.15\pm 0.14$  & $14.62\pm 0.12$  & $32.92\pm 0.17$  & $35.32\pm 0.15$ \\
& $4$  & $19.08\pm 0.11$  & $14.68\pm 0.12$  & $32.97\pm 0.08$  & $40.08\pm 0.1$ \\\hline
\multirow{ 5}{*}{2259} & $\frac{1}{4}$  & $18.24\pm 0.09$  & $17.11\pm 0.05$  & $8.56\pm 0.05$  & $16.22\pm 0.07$ \\
& $\frac{1}{2}$  & $24.75\pm 0.18$  & $23.74\pm 0.12$  & $17.34\pm 0.11$  & $25.15\pm 0.13$ \\
& $1$  & $24.77\pm 0.13$  & $33.27\pm 0.11$  & $34.12\pm 0.15$  & $36.44\pm 0.12$ \\
& $2$  & $24.68\pm 0.08$  & $43.09\pm 0.13$  & $42.93\pm 0.09$  & $50.49\pm 0.12$ \\
& $4$  & $24.76\pm 0.14$  & $43.29\pm 0.11$  & $42.54\pm 0.14$  & $53.2\pm 0.15$ \\\hline
\multirow{ 5}{*}{2261} & $\frac{1}{4}$  & $14.52\pm 0.14$  & $13.61\pm 0.1$  & $6.26\pm 0.05$  & $16.08\pm 0.08$ \\
& $\frac{1}{2}$  & $19.19\pm 0.1$  & $18.31\pm 0.09$  & $12.52\pm 0.08$  & $19.78\pm 0.05$ \\
& $1$  & $19.14\pm 0.13$  & $24.98\pm 0.12$  & $26.54\pm 0.09$  & $28.92\pm 0.09$ \\
& $2$  & $19.27\pm 0.14$  & $32.69\pm 0.09$  & $34.17\pm 0.08$  & $39.8\pm 0.13$ \\
& $4$  & $19.28\pm 0.12$  & $35.94\pm 0.08$  & $35.1\pm 0.13$  & $46.15\pm 0.12$ \\\hline
\multirow{ 5}{*}{2821} & $\frac{1}{4}$  & $5.89\pm 0.08$  & $4.31\pm 0.04$  & $3.47\pm 0.05$  & $6.58\pm 0.05$ \\
& $\frac{1}{2}$  & $8.19\pm 0.06$  & $5.84\pm 0.1$  & $6.62\pm 0.05$  & $9.68\pm 0.02$ \\
& $1$  & $8.18\pm 0.08$  & $8.4\pm 0.08$  & $13.21\pm 0.09$  & $13.6\pm 0.05$ \\
& $2$  & $8.19\pm 0.07$  & $11.99\pm 0.12$  & $14.18\pm 0.08$  & $19.3\pm 0.06$ \\
& $4$  & $8.2\pm 0.06$  & $15.82\pm 0.11$  & $16.04\pm 0.07$  & $23.3\pm 0.14$ \\\hline
\multirow{ 5}{*}{2997} & $\frac{1}{4}$  & $10.35\pm 0.08$  & $11.45\pm 0.1$  & $7.46\pm 0.08$  & $3.62\pm 0.05$ \\
& $\frac{1}{2}$  & $11.34\pm 0.09$  & $13.48\pm 0.08$  & $10.26\pm 0.1$  & $7.79\pm 0.07$ \\
& $1$  & $11.32\pm 0.06$  & $16.32\pm 0.12$  & $17.18\pm 0.12$  & $21.75\pm 0.14$ \\
& $2$  & $11.37\pm 0.09$  & $19.43\pm 0.09$  & $31.96\pm 0.12$  & $32.12\pm 0.1$ \\
& $4$  & $11.33\pm 0.1$  & $19.58\pm 0.1$  & $32.46\pm 0.12$  & $43.98\pm 0.1$ \\\hline
\multirow{ 5}{*}{3358} & $\frac{1}{4}$  & $18.92\pm 0.08$  & $17.74\pm 0.09$  & $14.91\pm 0.04$  & $19.4\pm 0.06$ \\
& $\frac{1}{2}$  & $27.31\pm 0.08$  & $24.74\pm 0.08$  & $23.7\pm 0.08$  & $28.0\pm 0.1$ \\
& $1$  & $27.22\pm 0.09$  & $33.65\pm 0.14$  & $36.08\pm 0.14$  & $38.14\pm 0.1$ \\
& $2$  & $27.19\pm 0.14$  & $42.57\pm 0.13$  & $35.2\pm 0.18$  & $50.8\pm 0.09$ \\
& $4$  & $27.29\pm 0.12$  & $43.83\pm 0.18$  & $36.12\pm 0.04$  & $64.34\pm 0.17$ \\\hline
\multirow{ 5}{*}{3386} & $\frac{1}{4}$  & $13.0\pm 0.1$  & $9.42\pm 0.06$  & $6.78\pm 0.08$  & $12.74\pm 0.12$ \\
& $\frac{1}{2}$  & $18.53\pm 0.1$  & $11.8\pm 0.07$  & $13.37\pm 0.08$  & $18.96\pm 0.13$ \\
& $1$  & $20.66\pm 0.09$  & $14.02\pm 0.07$  & $24.15\pm 0.1$  & $26.38\pm 0.12$ \\
& $2$  & $20.68\pm 0.1$  & $14.1\pm 0.13$  & $29.02\pm 0.09$  & $32.97\pm 0.09$ \\
& $4$  & $20.61\pm 0.13$  & $14.2\pm 0.1$  & $33.64\pm 0.15$  & $34.72\pm 0.12$ \\\hline
\multirow{ 5}{*}{3427} & $\frac{1}{4}$  & $16.49\pm 0.11$  & $15.78\pm 0.12$  & $10.29\pm 0.05$  & $19.56\pm 0.11$ \\
& $\frac{1}{2}$  & $22.97\pm 0.12$  & $18.73\pm 0.1$  & $20.11\pm 0.12$  & $30.29\pm 0.09$ \\
& $1$  & $22.94\pm 0.12$  & $18.88\pm 0.09$  & $39.96\pm 0.07$  & $41.66\pm 0.08$ \\
& $2$  & $22.89\pm 0.14$  & $18.89\pm 0.08$  & $50.8\pm 0.16$  & $54.94\pm 0.12$ \\
& $4$  & $22.91\pm 0.19$  & $18.94\pm 0.1$  & $69.9\pm 0.12$  & $68.95\pm 0.14$ \\\hline
\multirow{ 5}{*}{3476} & $\frac{1}{4}$  & $17.27\pm 0.08$  & $14.9\pm 0.1$  & $9.32\pm 0.04$  & $16.82\pm 0.08$ \\
& $\frac{1}{2}$  & $26.26\pm 0.07$  & $20.58\pm 0.13$  & $18.52\pm 0.1$  & $26.74\pm 0.05$ \\
& $1$  & $27.47\pm 0.12$  & $26.1\pm 0.12$  & $36.46\pm 0.07$  & $37.8\pm 0.07$ \\
& $2$  & $27.4\pm 0.1$  & $27.58\pm 0.15$  & $41.0\pm 0.17$  & $48.78\pm 0.11$ \\
& $4$  & $27.36\pm 0.12$  & $27.58\pm 0.17$  & $41.25\pm 0.18$  & $65.08\pm 0.11$ \\\hline
        \end{tabular} 
    }
    
    \caption{Average rewards (Impression) in $\%$ over the trajectories of iPinYou advertiser}
    \label{tab:ipin-result}
\end{table}

\begin{table}[t]
    \centering
    \scalebox{.7}{
        \begin{tabular}{ll llll}
            \hline 
            Adv   & Budget &  LinBid & RLB  & FDQI  & Ex-DDQN  \\ 
            \hline
\multirow{ 5}{*}{Adobe} & $\frac{1}{4}$  & $0.81\pm 0.01$  & $3.8\pm 0.04$  & $1.8\pm 0.02$  & $3.66\pm 0.06$ \\
& $\frac{1}{2}$  & $1.61\pm 0.01$  & $5.03\pm 0.06$  & $3.47\pm 0.02$  & $5.24\pm 0.1$ \\
& $1$  & $3.24\pm 0.02$  & $6.48\pm 0.07$  & $5.88\pm 0.03$  & $7.55\pm 0.05$ \\
& $2$  & $6.47\pm 0.03$  & $7.28\pm 0.06$  & $7.72\pm 0.03$  & $11.42\pm 0.07$ \\
& $4$  & $12.89\pm 0.06$  & $7.31\pm 0.09$  & $17.61\pm 0.14$  & $18.18\pm 0.09$ \\\hline
        \end{tabular} 
    }
    \caption{Average rewards (Impression) in $\%$ over the trajectories on Adobe Dataset}
    \label{tab:adobe-result}
\end{table}

\section{Architecture and Hyper-parameters}
\subsection{Architecture for Maximum Entropy Market State Model} The generator (or sampler) for WGAN-GP has an input size of $64$. We sample from standard normal distribution $\CN(0, I_{64})$ as the input to the generator. We use three hidden layers with $256,256,128$ nodes respectively. The output layer has categorical constraints (for each field in the feature vector). We use the Gumbel-softmax trick with a temperature parameter of $\tau=0.667$ while training. 
The critic has also three hidden layers with $256,256,128$ nodes respectively. We use a single node in the last layer to generate the critic scalar output similar to WGAN-GP.
We use $\ell_2$ regularization with a value of $1e-10$. We use 5 critic iteration for each generator steps with $\lambda=10$ as the gradient penalty  \cite{wgan-gp}. We use a batch size of 1024. We use default Xavier initialization for the critic and the generator.

\subsection{Hyper-parameter for Maximum Entropy Market State Model} We used learning rate from $\{1e-4, 2e-4\}$ for Adam optimizer. We run the maximum entropy model for both training and test batch dataset for each advertiser and run till convergence of the critic. Critic convergence implies the same distribution of the sampler as the true distribution \cite{wgan,wgan-gp}. Although we set maximum iterations to 4000, critic converges within 100-2000 iterations for all advertisers.

\subsection{Impression and Click Model} Impression model is a simple linear model with the input connected to the two-node layer to predict the expected market price and expected log variance of the market price distribution. We set initial weight parameters to $0$ and the bias parameter to $200$ and $10$ for mean and log variance parameter based on the empirical dataset.
The click model is a simple logistic regression initialized from the standard normal distribution.

\subsection{Hyper-parameter for Impression and Click Model}
We use $\ell_2$ regularization from $\{1e-2, 1e-4, 1e-6,1e-8\}$  for both impression and click model to do early stopping on the validation dataset. We use batchsize of $1024$ and run $100$ ($\times \mbox{number-of-batch}$) iteration.

\subsection{Ex-DDQN Architecture \& Hyper-parameters}
We use a neural network to learn the agent. We use the bid request, time constraint and budget constraint as the input to the Q-network. We connect the bid request to a single node hidden layer ($f(\bx)=h_1\in R^1$) such that a large number of features from bid requests do not overpower the budget and time constraints. We concatenate the single node hidden layer with budget and time constraints to compute the second layer hidden nodes as $h_2= [f_1(\bx)\oplus\bb\oplus \bt]$ which passes through the remaining neural network. We use one shared layer of dimension $128$ and followed by two separate networks to compute the value function and advantage function. The value function network has two-layer with $64$ and $1$ hidden nodes. The advantage function network has similarly a layer with a $64$ node followed by $20$ nodes. We quantize the bid values into  $20$ same length intervals to represent the action space. For the first step $f_1$, we initialize the network with the impression model weight parameters. We initialize the rest of the neural network using Xavier's initialization.

We use $\epsilon$-greedy with epsilon decay to sample from the action space. We use the following formula at time step $t$; $\epsilon = 0.2 + (1-0.2) \exp(-t/500,000)$. We use the memory buffer with a maximum length of $2,500,000$. We copy the target parameter every $5000$ time steps and start optimizing from time step $2000$. We run $5$ million steps with $16$ workers synchronously for interacting with the environment. We set the starting budget as $\alpha\times \mbox{cpm}_{train}\times 10^{-3}\times \mbox{timesteps}$ where $\alpha$ was randomly drawn from $\CU(-2,2)$, $\CU$ being the uniform distribution. The starting time constraint is always set to $100,000$.
We use learning rate parameters from $\{1e-3,2e-4\}$ for hyper-parameter tuning.

\subsection{FDQI Architecture \& Hyper-parameters}  For iPinYou, the batch dataset does not have a state (such as budget and time constraints) associated with the advertiser. We divide the batch dataset into multiple sequences of $100,000$ consecutive steps. We set the initial time constraint as $\bt= 100,000$e and set the initial budget constraint as the cost to be incurred by the advertiser $\bb = \sum_{i=1}^{100000} \mbox{cost}_i$ . 
We use the same architecture as Ex-DDQN for the value function estimator in Fitted Deep Q-Network. We do not have any memory buffer restriction; we use all the batch training interactions in the memory buffer.
Similar to Ex-DDQN, we use the learning rate parameter from $\{1e-3,2e-4\}$ to pick the best agent from the batch datasets.

\subsection{Test Procedure in Simulated Environment} We use the train batch dataset to learn the train market state model and the impression model. We use these two models to learn our Ex-DDQN agent. For all models, we use the same testing simulated environments for which we use the test batch dataset to learn the test market state model and the test impression model. These two models allow us to sample bid requests, and the market price distribution, necessary to evaluate any bidding algorithm. In no cases, the testing environment and the training environment share any samples to learn from. We divided train, test, validation split based on days which results in significant co-variates shift.


\balance